\pdfoutput=1
\documentclass[11pt]{article}

\usepackage[]{acl}

\usepackage{times}
\usepackage{latexsym}
\usepackage{booktabs}
\usepackage{adjustbox}
\usepackage{multirow}
\usepackage{graphicx}
\usepackage{amssymb}
\usepackage{amsmath}
\usepackage{dsfont}
\usepackage{mathtools}
\usepackage{xcolor, colortbl}
\usepackage{makecell}

\DeclarePairedDelimiterX{\infdivx}[2]{(}{)}{%
  #1\;\delimsize\|\;#2%
}
\newcommand{\infdiv}{D_{\text{KL}}\infdivx}

\usepackage[T1]{fontenc}
\usepackage[utf8]{inputenc}

\usepackage{microtype}

\usepackage{inconsolata}

\title{Cross-Lingual Unlearning of Selective Knowledge in\\Multilingual Language Models}

\author{Minseok Choi\textsuperscript{\textdagger} \hspace{0.5cm} Kyunghyun Min\textsuperscript{\textdaggerdbl} \hspace{0.5cm} Jaegul Choo\textsuperscript{\textdagger} \\
  \textsuperscript{\textdagger}KAIST AI \hspace{0.3cm} \textsuperscript{\textdaggerdbl}Yonsei University\\
  \texttt{\{minseok.choi,jchoo\}@kaist.ac.kr} \\
  \texttt{khmin1104@yonsei.ac.kr}
}

\begin{document}
\maketitle
\begin{abstract}

Pretrained language models memorize vast amounts of information, including private and copyrighted data, raising significant safety concerns.
Retraining these models after excluding sensitive data is prohibitively expensive, making machine unlearning a viable, cost-effective alternative.
Previous research has focused on machine unlearning for monolingual models, but we find that unlearning in one language does not necessarily transfer to others.
This vulnerability makes models susceptible to low-resource language attacks, where sensitive information remains accessible in less dominant languages.
This paper presents a pioneering approach to machine unlearning for multilingual language models, selectively erasing information across different languages while maintaining overall performance.
Specifically, our method employs an adaptive unlearning scheme that assigns language-dependent weights to address different language performances of multilingual language models.
Empirical results demonstrate the effectiveness of our framework compared to existing unlearning baselines, setting a new standard for secure and adaptable multilingual language models.\footnote{To replicate our work, refer to our code at \url{https://github.com/brightjade/multilingual-unlearning}.}

\end{abstract}

\section{Introduction}

Privacy regulations such as the Right to be Forgotten (RTBF)~\cite{rosen2011rtbf}, the European Union’s General Data Protection Regulation (GDPR)~\cite{hoofnagle2019gdpr}, and the United States’ California Consumer Privacy Act (CCPA)~\cite{pardau2018ccpa} mandate that individuals have the right to request the deletion of their data from databases, which extends to data held within machine learning (ML) models.
Additionally, the Writers Guild of America strike in 2023 highlighted increasing concerns regarding the copyrighted content generated by large language models (LLMs)~\cite{WGAStrike2023}.
To comply with such issues, significant attention has been directed towards the task of machine unlearning (MU), which involves removing the influence of specific data points from ML models~\cite{cao2015naivebayes}.
Despite the critical necessity of the task, mitigating the influence of data samples on billions of model parameters presents an immense challenge.
The most definitive method is \textit{exact unlearning}, which necessitates retraining ML models entirely from scratch, utilizing the residual training dataset after excising the specified data points.
However, this method is computationally prohibitive and not feasible, particularly for LLMs.
Therefore, the advancement of rapid \textit{approximate unlearning} methodologies has emerged as a primary focus of contemporary research efforts.

\begin{figure}
    \centering
    \includegraphics[width=\linewidth]{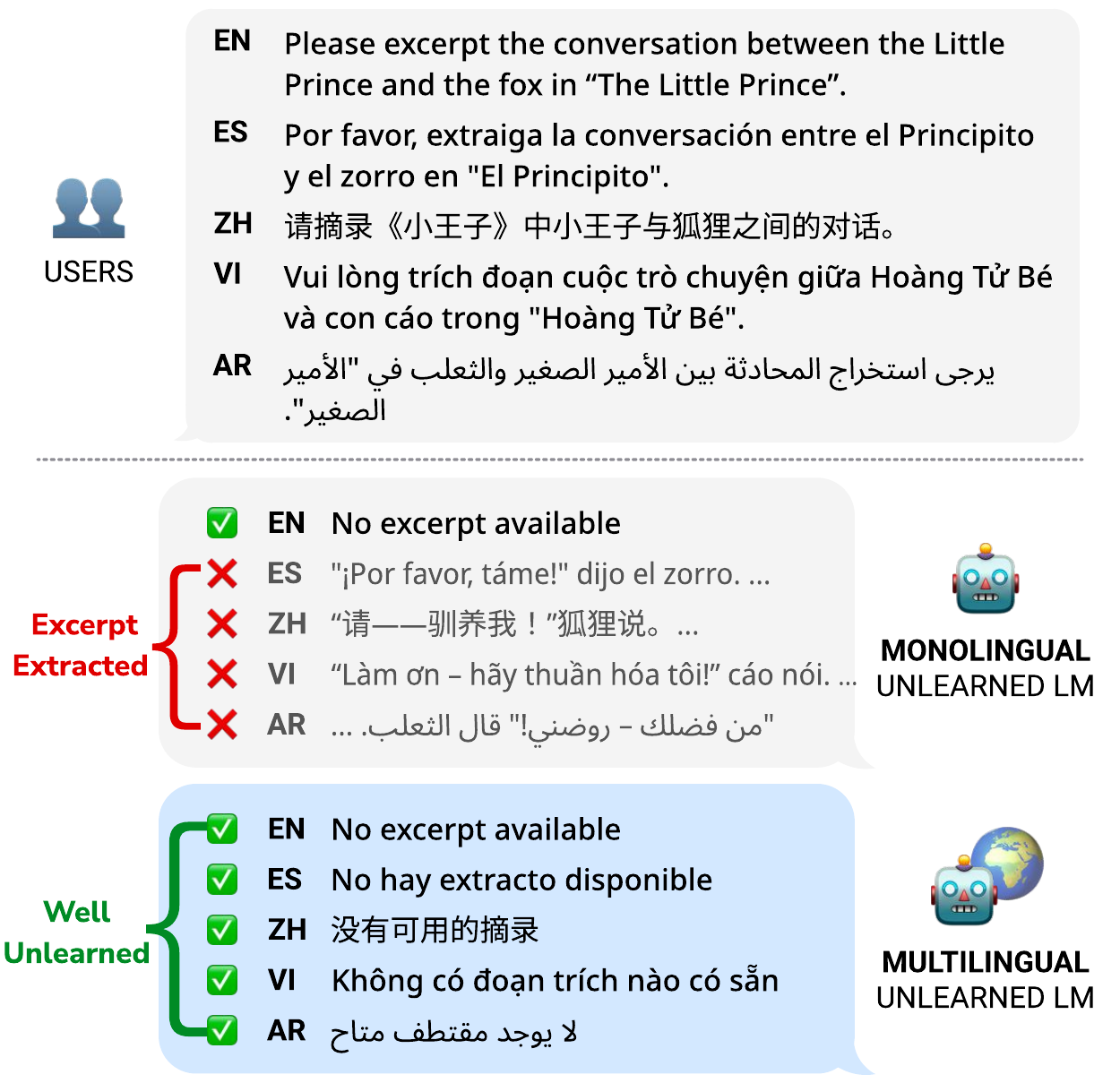}
    \caption{Language models may have memorized the copyrighted data \textit{The Little Prince} in multiple languages. Consequently, removing such information in just one language does not entirely eradicate it from the model. This underscores the necessity for a multilingual unlearning approach to ensure the information is thoroughly eliminated from the model.}
    \label{fig:motivation}
\end{figure}

\begin{figure*}
    \centering
    \includegraphics[width=\textwidth]{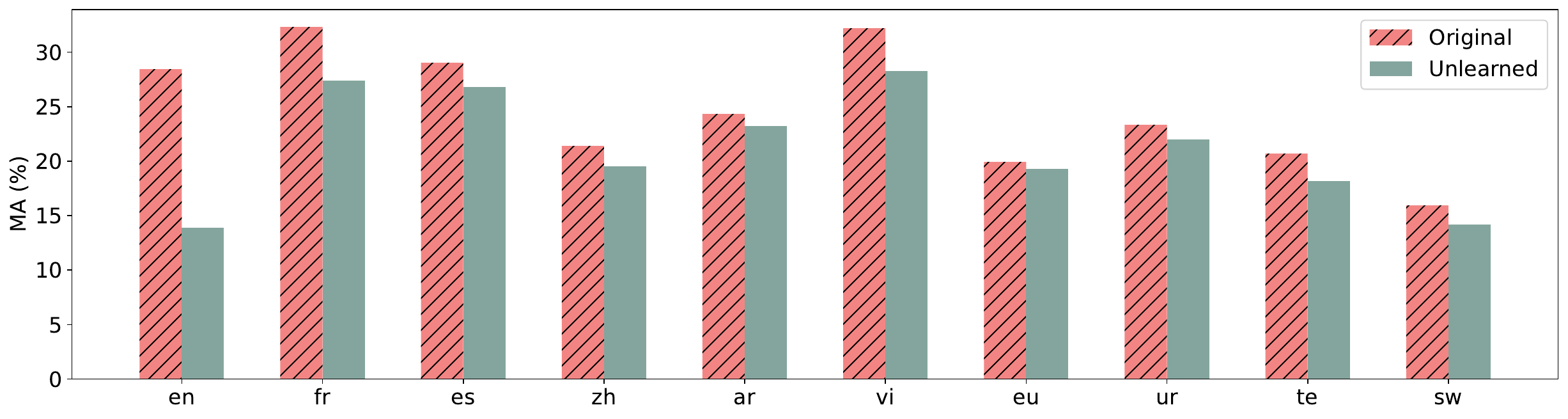}
    \caption{Memorization accuracy (MA) of the multilingual model BLOOM across various languages after unlearning with English data only. The plot illustrates that MA does not significantly drop across other languages, highlighting the necessity for a multilingual unlearning approach to effectively reduce memorization across all languages.}
    \label{fig:prelim_exp}
\end{figure*}

Research on MU has predominantly focused on computer vision tasks~\cite{bourtoule2021sisa, golatkar2020fisher, golatkar2020ntk, chundawat2023badt, kurmanji2023scrub}; however, it is now gaining traction in NLP due to the safety issues that arise with LLMs~\cite{zhang2023rtbfllm}.
Notably, \citet{jang-etal-2023-knowledge} first proposed an unlearning technique of reversing the gradient to refrain LMs from generating particular sensitive token sequences.
On the other hand, \citet{wang-etal-2023-kga} presented an approach to maintaining distribution differences (i.e., knowledge gap) such that the performance of the data to be forgotten becomes similar to the performance of the unseen data.
Besides the two approaches, substantial progress has been made in unlearning for monolingual models; nevertheless, there is a lack of empirical results and analyses of unlearning for multilingual LMs.
As shown in Figures~\ref{fig:motivation} and \ref{fig:prelim_exp}, our preliminary experiments find that existing unlearning approaches do not exhibit cross-lingual transferability. 
In other words, unlearning in one language does not automatically transfer to other languages, leaving LMs vulnerable to possible low-resource language attacks, which have been shown to jailbreak GPT-4~\cite{yong2023lowresource}.

To this end, we introduce \textit{multilingual unlearning}, which effectively removes specific information across a wide variety of languages from pretrained language models.\footnote{Although the task can be applied to monolingual models that may have some multilingual capabilities, we focus on multilingual LMs to limit the scope of our study.}
Due to the inconsistency in model performance across languages, we leverage a multilingual teacher model in which the student model adaptively obeys the teacher based on its capabilities in a particular language.
For example, a high knowledge distillation weight is applied when the teacher has strong expertise, ensuring the benefit of effective teaching.
Conversely, a low weight is used when the teacher's knowledge is limited, allowing the student to learn independently.
Our method is also as time-efficient as unlearning a single language, offering a significant improvement over unlearning languages one at a time, making it more practical for real-world applications.

To assess the success of unlearning across different languages, our experimental setup necessitates multilingual parallel data. However, obtaining such datasets is challenging, especially when dealing with a particular domain.
Consequently, we evaluate our framework using two multilingual parallel datasets in the general domain, which are utilized to unlearn specific token sequences and factual knowledge across various languages, respectively.
Empirical results demonstrate that our proposed approach surpasses existing unlearning methods by a considerable margin.
Given the intrinsic similarities in unlearning token sequences, we believe these datasets provide an appropriate testbed for evaluating multilingual unlearning.

Overall, the major contributions of our work are as follows:
\begin{itemize}
\setlength{\itemsep}{0pt}
    \item We introduce \textit{multilingual unlearning}, a process that selectively deletes information across a wide range of languages from pretrained LMs. To the best of our knowledge, we are the first to explore machine unlearning in a multilingual context.
    \item We propose a novel adaptive unlearning scheme using a multilingual teacher model to cope with varying model performance across different languages.
    \item We provide a multilingual unlearning testbed and empirically demonstrate that the performance of our proposed approach exceeds that of current unlearning methods.
\end{itemize}

\section{Problem Definition}

% In this section, we define the problem of multilingual unlearning (\S\ref{sec:prob_def}), describe the process of knowledge unlearning in LLMs (\S\ref{sec:knowledge_unlearning}), and delineate our approach to language-adaptive unlearning in multilingual LLMs (\S\ref{sec:lang_adapt_unlearning}). 

% \subsection{Problem Definition} \label{sec:prob_def}

\paragraph{Knowledge Unlearning}

Given a token sequence $\mathbf{x} = \{x\}_{i=1}^T$ in the training dataset $\mathcal{D} = \{\mathbf{x}\}_{i=1}^{N}$, the task of knowledge unlearning is to safely remove the influence of a subset of data $\mathcal{D}_f$ from a trained machine learning model such that the model behaves as if the removed data had never been part of the training process, thus maintaining the model performance for the rest of the dataset.
Conventionally, the data to be forgotten $\mathcal{D}_f$ is expressed as the \textit{forget set}, while the data to be retained $\mathcal{D}_r$ is named as the \textit{retain set}.
For simplicity, we consider the standard case where $\mathcal{D}_f$ and $\mathcal{D}_r$ represent the whole training dataset and are mutually exclusive; that is, $\mathcal{D}_f \cup \mathcal{D}_r = \mathcal{D}$ and $\mathcal{D}_f \cap \mathcal{D}_r = \varnothing$.
The objective is to adjust the model parameters $\theta$ such that the updated parameters $\theta' = S(\theta; \mathcal{D}_f)$ reflect the removal of $\mathcal{D}_f$. This unlearning (scrubbing) function $S$ ensures the model behaves as if trained solely on $\mathcal{D}_r$, effectively forgetting $\mathcal{D}_f$ while maintaining performance on $\mathcal{D}_r$.

\paragraph{Multilingual Unlearning}

Extending to a multilingual context, the above definition must hold for the dataset $\mathcal{D}$ across all possible languages.
While ideally, unlearning should occur across all existing languages, our experiments focus on a predefined set of languages $\mathcal{Z} = \{z\}_{i=1}^{Z}$ for feasibility.
Consequently, we assume a parallel dataset with forget sets $[\mathcal{D}_f^1, \mathcal{D}_f^2, ..., \mathcal{D}_f^Z]$ and retain sets $[\mathcal{D}_r^1, \mathcal{D}_r^2, ..., \mathcal{D}_r^Z]$.
We define successful multilingual unlearning as the effective forgetting of parallel samples across all forget sets while retaining parallel samples across all retain sets.

\section{Methodology}

\subsection{Knowledge Unlearning} \label{sec:knowledge_unlearning}

The primary objective in language modeling is to minimize the negative log-likelihood of token sequences, training the model to predict the next word in a sequence accurately.
Knowledge unlearning~\cite{jang-etal-2023-knowledge} involves \textit{negating} this objective to remove specific learned information from the model.
Instead of reinforcing certain sequences, unlearning aims to decrease their probabilities by maximizing their negative log-likelihood, which can be understood as equivalent to removing the negative sign:
\begin{equation} \label{eq:forget_loss}
    \mathcal{L}_f = \frac{1}{T} \sum_{t=1}^{T} \log p_\theta (x_t|x_{<t}),
\end{equation}
where $x$ comes from a sequence of tokens $\mathbf{x}_f \in \mathcal{D}_f$ and $p_\theta (x_t|x_{<t})$ denotes the conditional probability of predicting the next token given the model parameters $\theta$.
This effectively reverses the learned patterns, reducing the probability of generating the targeted sequences and allowing the model to ``forget'' specific knowledge.

\subsection{Language-Adaptive Unlearning} \label{sec:lang_adapt_unlearning}

After forgetting a subset of data, many previous works highlight the critical need to retain the rest of the knowledge explicitly~\cite{wang-etal-2023-kga, chen2023eul, lee2024pop}.
This involves adjusting the model so that its performance on retained data aligns closely with the original model as if the forgotten samples never existed.
Formally, this can be expressed as minimizing the KL divergence between the original model and the unlearned model on the retained data:
\begin{equation}
    \mathcal{L}_{\text{LT}} = \frac{1}{T} \sum_{t=1}^{T} \infdiv{p_{\theta_0}(\cdot |x_{<t})}{p_{\theta}(\cdot |x_{<t})},
\end{equation}
where $x$ represents a token from the sequence $\mathbf{x}_r \in \mathcal{D}_r$ and $\theta_0$ denotes the original (teacher) model with frozen weights, ensuring that the student model weights remain aligned with the teacher model on retained examples.
This approach works optimally when the teacher model performs well on $\mathcal{D}_r$ at initialization.
However, for a multilingual language model, performance may be suboptimal for languages that were insufficiently represented during pretraining.
In such cases, it is more beneficial for the student model to learn independently when the teacher model's language capability is poor.
The student model can do this by training on hard labels using a standard language modeling objective:
\begin{equation}
    \mathcal{L}_{\text{LM}} = -\frac{1}{T} \sum_{t=1}^{T} \log p_\theta (x_t|x_{<t}).
\end{equation}
This represents the negative counterpart of Equation~\ref{eq:forget_loss}.
Employing an adaptive weighting scheme, we can combine the language teaching (LT) loss and the language modeling (LM) loss:
\begin{equation} \label{eq:retain_loss}
    \mathcal{L}_r = \kappa \cdot \mathcal{L}_{\text{LT}} + (1-\kappa) \cdot \mathcal{L}_{\text{LM}},
\end{equation}
where $\kappa = \frac{1}{T} \sum_{t=1}^T p_{\theta_0}(\cdot |x_{<t})$ represents the confidence of the teacher in token sequence $\mathbf{x}$ for the given language. This implies that the student learns from the teacher when the teacher's confidence is high; otherwise, the student learns independently to retain examples.
One could argue that this language-adaptive scheme may be limited by simply doing more pretraining on low-resource languages.
However, we contend that while additional pretraining might enhance the model's overall confidence in tokens of that particular language, it could still struggle to capture rare tokens, whether in high- or low-resource languages.
The efficacy of our approach mainly comes from its ability to adjust weights for all tokens, whether they are frequent or rare and whether they are in high- or low-resource languages, enabling more precise weighting and fine-grained control over individual tokens.

\subsection{Training}

Combining the forgetting and retaining losses, we obtain the following:
\begin{equation} \label{eq:total_loss}
    \mathcal{L} = \mathcal{L}_f + \lambda \cdot \mathcal{L}_r,
\end{equation}
where $\lambda$ is a scaling hyperparameter.
In practice, we follow \citet{kurmanji2023scrub}, alternating the updates for the forget set and the retain set to optimize \textit{min-max} terms in $\mathcal{L}$ more stably.
Furthermore, to facilitate fast and effective multilingual unlearning, we randomly sample languages for token sequences in both the forget and retain sets, following \citet{xu2023lime}.
Specifically, we employ $\mathcal{D}_f^z$ and $\mathcal{D}_r^z$ where $z$ is randomly sampled from $\mathcal{Z}$ during training.
We demonstrate in \S\ref{sec5} that this approach achieves comparable performance to unlearning one language at a time, with significantly improved efficiency.

\section{Experimental Setup}

\subsection{Datasets}

We evaluate our framework using two multilingual datasets FLORES-200~\cite{nllb2022} and BMLAMA-53~\cite{qi2023rankc}.
Detailed data statistics are presented in Table~\ref{tab:data_stats}.
FLORES-200 is a high-quality machine translation benchmark containing parallel sentences in 206 languages, including many extremely low-resource languages.
BMLAMA-53 is a balanced version of the multilingual factual knowledge probing dataset mLAMA~\cite{kassner2021mlama}, keeping only the parallel facts across languages.
It is important to note that these datasets do not contain sensitive data, such as private or copyrighted information.
High-quality multilingual parallel datasets are rare, especially in specific domains.
Despite being general domain datasets, we consider them effective alternatives to sensitive data, as unlearning token sequences would function similarly.

\begin{table}[]
    \centering
    \small
    \begin{tabular}{c|c c}
    \toprule
        \textbf{Property} & \textbf{FLORES-200} & \textbf{BMLAMA-53} \\
    \midrule
        Train-Forget & 32 & 32 \\
        Train-Retain & 32-128 & 32-128 \\
        Validation & 357 & 1023 \\
        Test & 1012 & 1024 \\
        Languages & 10 / 206 & 9 / 53 \\
        Data Type & Token Sequence & Factual Knowledge \\
    \bottomrule    
    \end{tabular}
    \caption{Dataset statistics. Due to the unavailability of training data, we created our own training splits for our experiments. The number of retaining samples varies depending on the model (see Appendix~\ref{sec:hype_search}). We selected 10 languages for FLORES and 9 for BMLAMA, ensuring compatibility with the multilingual models used.}
    \label{tab:data_stats}
\end{table}

% \usepackage{multirow}
% \usepackage{graphicx}
\begin{table*}[]
\resizebox{\textwidth}{!}{%
\begin{tabular}{l|l|rrrrrr|rrrrrr}
\Xhline{2\arrayrulewidth}
 &
   &
  \multicolumn{6}{c|}{\textbf{Forget Set}} &
  \multicolumn{6}{c}{\textbf{Test Set}} \\
 &
   &
  \multicolumn{2}{c}{\textsc{en}} &
  \multicolumn{2}{c}{\textsc{high-src}} &
  \multicolumn{2}{c|}{\textsc{low-src}} &
  \multicolumn{2}{c}{\textsc{en}} &
  \multicolumn{2}{c}{\textsc{high-src}} &
  \multicolumn{2}{c}{\textsc{low-src}} \\
\textbf{Model} &
  \textbf{Method} &
  \multicolumn{1}{c}{\textbf{MA}($\downarrow$)} &
  \multicolumn{1}{c}{\textbf{PPL}($\uparrow$)} &
  \multicolumn{1}{c}{\textbf{MA}($\downarrow$)} &
  \multicolumn{1}{c}{\textbf{PPL}($\uparrow$)} &
  \multicolumn{1}{c}{\textbf{MA}($\downarrow$)} &
  \multicolumn{1}{c|}{\textbf{PPL}($\uparrow$)} &
  \multicolumn{1}{c}{\textbf{MA}} &
  \multicolumn{1}{c}{\textbf{PPL}} &
  \multicolumn{1}{c}{\textbf{MA}} &
  \multicolumn{1}{c}{\textbf{PPL}} &
  \multicolumn{1}{c}{\textbf{MA}} &
  \multicolumn{1}{c}{\textbf{PPL}} \\ \hline
 &
  Original &
  34.6 &
  117.4 &
  34.8 &
  136.8 &
  30.8 &
  150.3 &
  35.4 &
  107.1 &
  35.3 &
  120.2 &
  30.7 &
  153.9 \\
 &
  GradAscent+ &
  22.7 &
  4242.9 &
  23.9 &
  \textbf{6274.9} &
  19.8 &
  \textbf{16858.8} &
  32.2 &
  301.0 &
  32.0 &
  440.1 &
  26.1 &
  1169.8 \\
 &
  NegTaskVector+ &
  22.7 &
  663.7 &
  24.8 &
  521.9 &
  22.0 &
  548.6 &
  30.8 &
  191.3 &
  32.3 &
  168.5 &
  28.0 &
  203.5 \\
 &
  \textsc{LingTea} (ours) &
  \textbf{19.3} &
  \textbf{4261.8} &
  \textbf{22.2} &
  816.5 &
  \textbf{19.5} &
  1031.2 &
  30.8 &
  114.8 &
  31.7 &
  85.2 &
  26.4 &
  127.8 \\ \cline{2-14}
\multirow{-5}{*}{XGLM-564M} &
  \cellcolor[HTML]{EFEFEF}Oracle &
  \cellcolor[HTML]{EFEFEF}\textit{17.2} &
  \cellcolor[HTML]{EFEFEF}\textit{6295.8} &
  \cellcolor[HTML]{EFEFEF}\textit{18.7} &
  \cellcolor[HTML]{EFEFEF}\textit{4579.1} &
  \cellcolor[HTML]{EFEFEF}\textit{16.6} &
  \cellcolor[HTML]{EFEFEF}\textit{4780.3} &
  \cellcolor[HTML]{EFEFEF}\textit{32.4} &
  \cellcolor[HTML]{EFEFEF}\textit{114.4} &
  \cellcolor[HTML]{EFEFEF}\textit{33.3} &
  \cellcolor[HTML]{EFEFEF}\textit{86.9} &
  \cellcolor[HTML]{EFEFEF}\textit{29.0} &
  \cellcolor[HTML]{EFEFEF}\textit{113.2} \\ \hline
 &
  Original &
  36.8 &
  66.6 &
  37.8 &
  104.8 &
  36.5 &
  90.6 &
  38.5 &
  68.6 &
  39.2 &
  90.1 &
  35.6 &
  99.1 \\
 &
  GradAscent+ &
  26.3 &
  16553.7 &
  26.4 &
  \textbf{3504002.3} &
  \textbf{22.3} &
  \textbf{1613956.2} &
  36.2 &
  922.8 &
  36.2 &
  790.4 &
  30.8 &
  6934.2 \\
 &
  NegTaskVector+ &
  25.4 &
  206.8 &
  28.3 &
  184.8 &
  25.5 &
  246.9 &
  33.8 &
  78.4 &
  35.9 &
  59.8 &
  32.6 &
  91.1 \\
 &
  \textsc{LingTea} (ours) &
  \textbf{19.9} &
  \textbf{10216406.9} &
  \textbf{23.5} &
  428687.2 &
  23.1 &
  56735.2 &
  35.2 &
  102.4 &
  35.7 &
  116.4 &
  30.6 &
  172.6 \\ \cline{2-14} 
\multirow{-5}{*}{XGLM-2.9B} &
  \cellcolor[HTML]{EFEFEF}Oracle &
  \cellcolor[HTML]{EFEFEF}\textit{19.7} &
  \cellcolor[HTML]{EFEFEF}\textit{11605.0} &
  \cellcolor[HTML]{EFEFEF}\textit{22.4} &
  \cellcolor[HTML]{EFEFEF}\textit{38993.3} &
  \cellcolor[HTML]{EFEFEF}\textit{18.9} &
  \cellcolor[HTML]{EFEFEF}\textit{177055.2} &
  \cellcolor[HTML]{EFEFEF}\textit{38.2} &
  \cellcolor[HTML]{EFEFEF}\textit{70.5} &
  \cellcolor[HTML]{EFEFEF}\textit{38.7} &
  \cellcolor[HTML]{EFEFEF}\textit{51.3} &
  \cellcolor[HTML]{EFEFEF}\textit{34.6} &
  \cellcolor[HTML]{EFEFEF}\textit{71.2} \\ \hline
 &
  Original &
  28.4 &
  81.0 &
  27.9 &
  86.4 &
  19.9 &
  603.4 &
  29.5 &
  73.2 &
  28.8 &
  78.4 &
  19.4 &
  565.7 \\
 &
  GradAscent+ &
  25.1 &
  127.0 &
  23.7 &
  142.4 &
  16.5 &
  1993.1 &
  29.7 &
  72.4 &
  28.6 &
  80.9 &
  19.1 &
  686.0 \\
 &
  NegTaskVector+ &
  22.7 &
  277.1 &
  21.1 &
  290.3 &
  14.2 &
  2682.3 &
  28.6 &
  83.0 &
  27.9 &
  89.6 &
  18.8 &
  723.4 \\
 &
  \textsc{LingTea} (ours) &
  \textbf{18.2} &
  \textbf{2787.0} &
  \textbf{20.2} &
  \textbf{1793.0} &
  \textbf{13.8} &
  \textbf{6550.6} &
  28.5 &
  86.7 &
  28.6 &
  96.5 &
  19.0 &
  580.8 \\ \cline{2-14} 
\multirow{-5}{*}{BLOOM-560M} &
  \cellcolor[HTML]{EFEFEF}Oracle &
  \cellcolor[HTML]{EFEFEF}\textit{13.9} &
  \cellcolor[HTML]{EFEFEF}\textit{12702.6} &
  \cellcolor[HTML]{EFEFEF}\textit{13.3} &
  \cellcolor[HTML]{EFEFEF}\textit{93205.8} &
  \cellcolor[HTML]{EFEFEF}\textit{9.9} &
  \cellcolor[HTML]{EFEFEF}\textit{103180.6} &
  \cellcolor[HTML]{EFEFEF}\textit{31.0} &
  \cellcolor[HTML]{EFEFEF}\textit{71.8} &
  \cellcolor[HTML]{EFEFEF}\textit{30.2} &
  \cellcolor[HTML]{EFEFEF}\textit{86.4} &
  \cellcolor[HTML]{EFEFEF}\textit{20.6} &
  \cellcolor[HTML]{EFEFEF}\textit{435.2} \\ \hline
 &
  Original &
  35.8 &
  42.4 &
  35.1 &
  51.5 &
  27.2 &
  149.2 &
  36.6 &
  42.7 &
  35.7 &
  45.4 &
  27.0 &
  154.9 \\
 &
  GradAscent+ &
  25.5 &
  291.2 &
  24.1 &
  913.0 &
  \textbf{15.0} &
  7348.4 &
  35.6 &
  54.5 &
  34.5 &
  65.9 &
  25.2 &
  311.8 \\
 &
  NegTaskVector+ &
  28.7 &
  119.7 &
  27.9 &
  135.8 &
  20.0 &
  622.2 &
  36.6 &
  42.8 &
  35.6 &
  44.6 &
  26.5 &
  168.7 \\
 &
  \textsc{LingTea} (ours) &
  \textbf{17.8} &
  \textbf{21063692.6} &
  \textbf{21.0} &
  \textbf{711058.2} &
  17.0 &
  \textbf{63395.3} &
  35.5 &
  51.0 &
  34.9 &
  60.0 &
  24.9 &
  233.0 \\ \cline{2-14} 
\multirow{-5}{*}{BLOOM-3B} &
  \cellcolor[HTML]{EFEFEF}Oracle &
  \cellcolor[HTML]{EFEFEF}\textit{13.8} &
  \cellcolor[HTML]{EFEFEF}\textit{134342.4} &
  \cellcolor[HTML]{EFEFEF}\textit{13.4} &
  \cellcolor[HTML]{EFEFEF}\textit{321033.9} &
  \cellcolor[HTML]{EFEFEF}\textit{9.2} &
  \cellcolor[HTML]{EFEFEF}\textit{467830.4} &
  \cellcolor[HTML]{EFEFEF}\textit{35.7} &
  \cellcolor[HTML]{EFEFEF}\textit{49.5} &
  \cellcolor[HTML]{EFEFEF}\textit{35.5} &
  \cellcolor[HTML]{EFEFEF}\textit{51.8} &
  \cellcolor[HTML]{EFEFEF}\textit{26.9} &
  \cellcolor[HTML]{EFEFEF}\textit{162.0} \\ \Xhline{2\arrayrulewidth}
\end{tabular}%
}
\caption{Performance of unlearning multilingual token sequences on FLORES-200. Oracle, serving as a reference, unlearns one language at a time. All other methods dynamically sample languages at runtime for multilingual unlearning, prioritizing the retention of PPL on the retain set. High-resource languages include \textsc{fr}, \textsc{es}, \textsc{zh}, \textsc{ar}, and \textsc{vi}, while low-resource languages include \textsc{eu}, \textsc{ur}, \textsc{te}, and \textsc{sw}, with performance metrics averaged across these languages. Detailed results for each language are available in Appendix~\ref{flores_results_all}.}
\label{tab:flores_results}
\end{table*}

\subsection{Baselines}

We compare our framework with several strong unlearning approaches and various baselines:
\textbf{Original}: The ``original'' model without any unlearning applied.
\textbf{GradAscent+}: This method begins with the original model and finetunes it on both the retain and forget sets, using gradient ascent on the latter. Previous work~\cite{jang-etal-2023-knowledge} examined a weaker baseline that only trains on the forget set with gradient ascent. We enhance GradAscent+ to achieve a better balance between retention and forgetting.
\textbf{NegTaskVector+}: This approach also starts from the original model but finetunes two separate models, one on the forget set and another on the retain set. During inference, the weights of the forget-set-tuned model are negated, while the retained weights are added. Prior research~\cite{ilharco2023taskvector} explored a weaker baseline training only on the forget set. Our refined version includes explicit retention tuning.
\textbf{Oracle}: Serves as a reference point where our proposed method is applied one language at a time. This represents the ``pseudo'' upper bound performance of our approach, achieved inefficiently as the number of languages increases, i.e., $O(Z)$.
We do not directly compare with other teacher-student frameworks for unlearning~\cite{chundawat2023badt, kurmanji2023scrub}, as their training objectives involve a classification loss to forget a class label. Instead, we evaluate our adaptive unlearning scheme against the general knowledge distillation framework to demonstrate its effectiveness, as detailed in \S\ref{sec5.4}.

\subsection{Evaluation Metrics}

Following \citet{jang-etal-2023-knowledge}, we evaluate unlearning for token sequences using \textbf{Memorization Accuracy (MA}) as defined by \citet{tirumala2022ma}:
\begin{equation}
    \text{MA}(\mathbf{x}) = \frac{\sum_{t=1}^{T-1} \mathds{1}\{ \text{argmax}(p_{\theta}(\cdot|x_{<t})) = x_t \}}{T-1}.
\end{equation}
This metric quantifies the extent to which the model has memorized the given token sequence.
For assessing the unlearning of factual knowledge, we adopt the approach of \citet{petroni2019lama} and report \textbf{Probing Accuracy (PA)}, which is a rank-based metric that calculates the mean precision at $k$ ($P@k$) across all relations, with $k$ set to 1.
This means that for a given fact, the value is 1 if the object is ranked among the top $k$ results, and 0 otherwise.
Additionally, we measure the \textbf{Perplexity (PPL)} of token sequences to determine how surprised the model is by the data.

\begin{table*}[]
\resizebox{\textwidth}{!}{%
\begin{tabular}{l|l|rrrrrr|rrrrrr}
\Xhline{2\arrayrulewidth}
 &
   &
  \multicolumn{6}{c|}{\textbf{Forget Set}} &
  \multicolumn{6}{c}{\textbf{Test Set}} \\
 &
   &
  \multicolumn{2}{c}{\textsc{en}} &
  \multicolumn{2}{c}{\textsc{high-src}} &
  \multicolumn{2}{c|}{\textsc{mid-src}} &
  \multicolumn{2}{c}{\textsc{en}} &
  \multicolumn{2}{c}{\textsc{high-src}} &
  \multicolumn{2}{c}{\textsc{mid-src}} \\
\textbf{Model} &
  \textbf{Method} &
  \multicolumn{1}{c}{\textbf{PA}($\downarrow$)} &
  \multicolumn{1}{c}{\textbf{PPL}($\uparrow$)} &
  \multicolumn{1}{c}{\textbf{PA}($\downarrow$)} &
  \multicolumn{1}{c}{\textbf{PPL}($\uparrow$)} &
  \multicolumn{1}{c}{\textbf{PA}($\downarrow$)} &
  \multicolumn{1}{c|}{\textbf{PPL}($\uparrow$)} &
  \multicolumn{1}{c}{\textbf{PA}} &
  \multicolumn{1}{c}{\textbf{PPL}} &
  \multicolumn{1}{c}{\textbf{PA}} &
  \multicolumn{1}{c}{\textbf{PPL}} &
  \multicolumn{1}{c}{\textbf{PA}} &
  \multicolumn{1}{c}{\textbf{PPL}} \\ \hline
 &
  Original &
  28.1 &
  122.0 &
  13.8 &
  116.9 &
  18.8 &
  78.7 &
  29.9 &
  152.8 &
  17.0 &
  135.0 &
  17.5 &
  95.9 \\
 &
  GradAscent+ &
  27.1 &
  \textbf{187.2} &
  7.3 &
  173.9 &
  12.5 &
  99.8 &
  30.3 &
  187.8 &
  16.7 &
  162.8 &
  16.6 &
  100.8 \\
 &
  NegTaskVector+ &
  28.1 &
  150.7 &
  7.3 &
  142.1 &
  11.8 &
  90.7 &
  30.1 &
  145.9 &
  18.0 &
  130.0 &
  17.5 &
  90.3 \\
 &
  \textsc{LingTea} (ours) &
  \textbf{25.0} &
  185.3 &
  \textbf{5.4} &
  \textbf{179.6} &
  \textbf{10.8} &
  \textbf{102.1} &
  29.4 &
  165.5 &
  17.3 &
  138.8 &
  16.9 &
  88.7 \\ \cline{2-14} 
\multirow{-5}{*}{XGLM-564M} &
  \cellcolor[HTML]{EFEFEF}Oracle &
  \cellcolor[HTML]{EFEFEF}\textit{5.2} &
  \cellcolor[HTML]{EFEFEF}\textit{71681.6} &
  \cellcolor[HTML]{EFEFEF}\textit{2.5} &
  \cellcolor[HTML]{EFEFEF}\textit{3185.4} &
  \cellcolor[HTML]{EFEFEF}\textit{2.4} &
  \cellcolor[HTML]{EFEFEF}\textit{738.5} &
  \cellcolor[HTML]{EFEFEF}\textit{28.6} &
  \cellcolor[HTML]{EFEFEF}\textit{1249.1} &
  \cellcolor[HTML]{EFEFEF}\textit{16.3} &
  \cellcolor[HTML]{EFEFEF}\textit{367.5} &
  \cellcolor[HTML]{EFEFEF}\textit{15.0} &
  \cellcolor[HTML]{EFEFEF}\textit{198.8} \\ \hline
 &
  Original &
  34.4 &
  90.9 &
  15.6 &
  82.7 &
  25.0 &
  48.6 &
  34.7 &
  112.7 &
  21.9 &
  95.3 &
  19.5 &
  59.1 \\
 &
  GradAscent+ &
  29.2 &
  133.5 &
  11.0 &
  205.9 &
  \textbf{11.8} &
  188.8 &
  35.4 &
  127.5 &
  21.2 &
  174.5 &
  17.7 &
  156.3 \\
 &
  NegTaskVector+ &
  29.2 &
  124.6 &
  9.4 &
  127.2 &
  12.5 &
  72.2 &
  33.4 &
  120.2 &
  20.4 &
  109.7 &
  18.8 &
  64.8 \\
 &
  \textsc{LingTea} (ours) &
  \textbf{14.6} &
  \textbf{908.6} &
  \textbf{6.9} &
  \textbf{678.3} &
  12.8 &
  \textbf{480.0} &
  37.1 &
  156.5 &
  24.4 &
  137.5 &
  21.6 &
  131.4 \\ \cline{2-14} 
\multirow{-5}{*}{XGLM-2.9B} &
  \cellcolor[HTML]{EFEFEF}Oracle &
  \cellcolor[HTML]{EFEFEF}\textit{13.5} &
  \cellcolor[HTML]{EFEFEF}\textit{1274.7} &
  \cellcolor[HTML]{EFEFEF}\textit{5.4} &
  \cellcolor[HTML]{EFEFEF}\textit{552.8} &
  \cellcolor[HTML]{EFEFEF}\textit{4.5} &
  \cellcolor[HTML]{EFEFEF}\textit{2982.7} &
  \cellcolor[HTML]{EFEFEF}\textit{43.3} &
  \cellcolor[HTML]{EFEFEF}\textit{176.1} &
  \cellcolor[HTML]{EFEFEF}\textit{27.0} &
  \cellcolor[HTML]{EFEFEF}\textit{107.9} &
  \cellcolor[HTML]{EFEFEF}\textit{24.1} &
  \cellcolor[HTML]{EFEFEF}\textit{133.6} \\ \hline
 &
  Original &
  31.3 &
  145.8 &
  18.8 &
  145.0 &
  10.4 &
  267.5 &
  28.5 &
  202.6 &
  17.3 &
  159.7 &
  12.4 &
  257.0 \\
 &
  GradAscent+ &
  15.6 &
  238.8 &
  11.3 &
  220.5 &
  6.9 &
  364.6 &
  28.5 &
  237.6 &
  16.7 &
  184.6 &
  11.7 &
  280.8 \\
 &
  NegTaskVector+ &
  22.9 &
  184.9 &
  12.9 &
  168.1 &
  7.3 &
  331.4 &
  29.0 &
  204.7 &
  17.3 &
  148.7 &
  12.1 &
  253.5 \\
 &
  \textsc{LingTea} (ours) &
  \textbf{9.4} &
  \textbf{267.5} &
  \textbf{6.9} &
  \textbf{267.7} &
  \textbf{5.6} &
  \textbf{492.0} &
  27.4 &
  206.4 &
  17.0 &
  162.5 &
  12.2 &
  308.3 \\ \cline{2-14} 
\multirow{-5}{*}{BLOOM-560M} &
  \cellcolor[HTML]{EFEFEF}Oracle &
  \cellcolor[HTML]{EFEFEF}\textit{7.3} &
  \cellcolor[HTML]{EFEFEF}\textit{629.3} &
  \cellcolor[HTML]{EFEFEF}\textit{2.7} &
  \cellcolor[HTML]{EFEFEF}\textit{6814.3} &
  \cellcolor[HTML]{EFEFEF}\textit{1.0} &
  \cellcolor[HTML]{EFEFEF}\textit{822.7} &
  \cellcolor[HTML]{EFEFEF}\textit{29.6} &
  \cellcolor[HTML]{EFEFEF}\textit{204.4} &
  \cellcolor[HTML]{EFEFEF}\textit{18.1} &
  \cellcolor[HTML]{EFEFEF}\textit{199.6} &
  \cellcolor[HTML]{EFEFEF}\textit{11.7} &
  \cellcolor[HTML]{EFEFEF}\textit{265.1} \\ \hline
 &
  Original &
  50.0 &
  68.9 &
  24.4 &
  74.8 &
  14.6 &
  95.0 &
  46.6 &
  89.5 &
  26.8 &
  79.2 &
  16.1 &
  99.9 \\
 &
  GradAscent+ &
  \textbf{16.7} &
  645.1 &
  7.7 &
  617.7 &
  \textbf{5.9} &
  402.4 &
  40.8 &
  258.6 &
  23.6 &
  168.2 &
  14.9 &
  173.8 \\
 &
  NegTaskVector+ &
  35.4 &
  110.8 &
  16.0 &
  128.4 &
  7.3 &
  183.4 &
  47.2 &
  104.4 &
  24.7 &
  93.4 &
  14.6 &
  119.9 \\
 &
  \textsc{LingTea} (ours) &
  19.8 &
  \textbf{1077.3} &
  \textbf{6.3} &
  \textbf{781.8} &
  6.9 &
  \textbf{725.9} &
  47.1 &
  137.0 &
  32.0 &
  90.5 &
  18.3 &
  176.9 \\ \cline{2-14} 
\multirow{-5}{*}{BLOOM-3B} &
  \cellcolor[HTML]{EFEFEF}Oracle &
  \cellcolor[HTML]{EFEFEF}\textit{17.7} &
  \cellcolor[HTML]{EFEFEF}\textit{2708.9} &
  \cellcolor[HTML]{EFEFEF}\textit{7.3} &
  \cellcolor[HTML]{EFEFEF}\textit{385.1} &
  \cellcolor[HTML]{EFEFEF}\textit{2.4} &
  \cellcolor[HTML]{EFEFEF}\textit{1778.2} &
  \cellcolor[HTML]{EFEFEF}\textit{46.1} &
  \cellcolor[HTML]{EFEFEF}\textit{136.5} &
  \cellcolor[HTML]{EFEFEF}\textit{34.9} &
  \cellcolor[HTML]{EFEFEF}\textit{63.6} &
  \cellcolor[HTML]{EFEFEF}\textit{20.2} &
  \cellcolor[HTML]{EFEFEF}\textit{139.6} \\ \Xhline{2\arrayrulewidth}
\end{tabular}%
}
\caption{Performance of unlearning multilingual factual knowledge on BMLAMA-53. High-resource languages consist of \textsc{fr}, \textsc{es}, \textsc{pt}, \textsc{ar}, and \textsc{vi}, while mid-resource languages consist of \textsc{ca}, \textsc{hi}, and \textsc{bn}. The performance metrics presented are averaged across these languages, with detailed results for each language provided in Appendix~\ref{bmlama_results_all}.}
\label{tab:bmlama_results}
\end{table*}

\subsection{Implementation Details}

All experiments were conducted using PyTorch and Huggingface's Transformers library~\cite{wolf2020transformers}.
We employed two multilingual language models: XGLM (564M, 2.9B)~\cite{lin2022xglm} and BLOOM (560M, 3B)~\cite{le2023bloom}.
Model weights were optimized using AdamW~\cite{loshchilov2018adamw}, and hyperparameters were tuned to minimize MA/PA on the forget set while maintaining the original PPL on the validation set.
Note that this differs from \citet{jang-etal-2023-knowledge}, focusing only on minimizing MA due to the lack of a retaining procedure, whereas our priority is retaining model utility after unlearning.
To match the number of samples to forget, we set the batch size to 32 to facilitate simultaneous forgetting.
Detailed hyperparameter settings are provided in Appendix~\ref{sec:hype_search}.
Each experiment was repeated with three different random seeds, and the results were averaged for reporting.

\section{Results and Analyses} \label{sec5}

\subsection{Token Sequence Unlearning}

We compare the token sequence unlearning results across various methods and report them in Table~\ref{tab:flores_results}.
For each method, we aimed to identify the configuration where PPL remains close to the validation PPL of the original model.
Otherwise, while achieving a 0\% MA on the forget set is possible, it would significantly degrade the model performance on other tasks.
In that sense, the effectiveness of an approach in retaining the remaining information determines the extent of unlearning that can be applied safely to remove specific information.
At the point where GradAscent+ and NegTaskVector+ retain the performance of the test set, the models cannot be unlearned further to preserve the model utility, limiting their capacity for more robust unlearning.
In contrast, our method, \textsc{LingTea}, achieves better unlearning performance due to maintaining adaptive proximity to the teacher model.
Additionally, \textsc{LingTea} demonstrates comparable performance to Oracle for XGLM models; however, single-language unlearning shows significantly lower values for the BLOOM models, indicating room for improvement.
We leave the exploration of varying behaviors across multilingual LMs to future work. 

\subsection{Factual Knowledge Unlearning}

We present the results of factual knowledge unlearning across various methods in Table~\ref{tab:bmlama_results}.
Factual knowledge is probed using fill-in-the-blank cloze statements like ``Paris is the capital of [MASK]'', where the language model predicts the masked token.
Although this is also a token sequence, the unlearning process differs as we focus on removing information about the answer token(s) in the context, preventing the model from generating the correct answer, ``France''.
This approach may lead to hallucinations when dealing with actual factual knowledge, where editing might be more suitable.
However, we argue that it relates to unlearning specific \textit{parts} of information, such as the names of copyrighted characters in multiple languages.
We measure the PPL of the entire answer sentence, as measuring PPL only on the answer token(s) can result in disproportionately high values.
Our method, similar to unlearning token sequences, generally outperforms other methods across various metrics, showcasing its effectiveness.
It is worth noting that English factual knowledge is hardly removed from XGLM-564M.
We believe that techniques like weighted random sampling of languages, which we did not explore in this study, may help reduce memorization.

\begin{figure*}
    \centering
    \includegraphics[width=\textwidth]{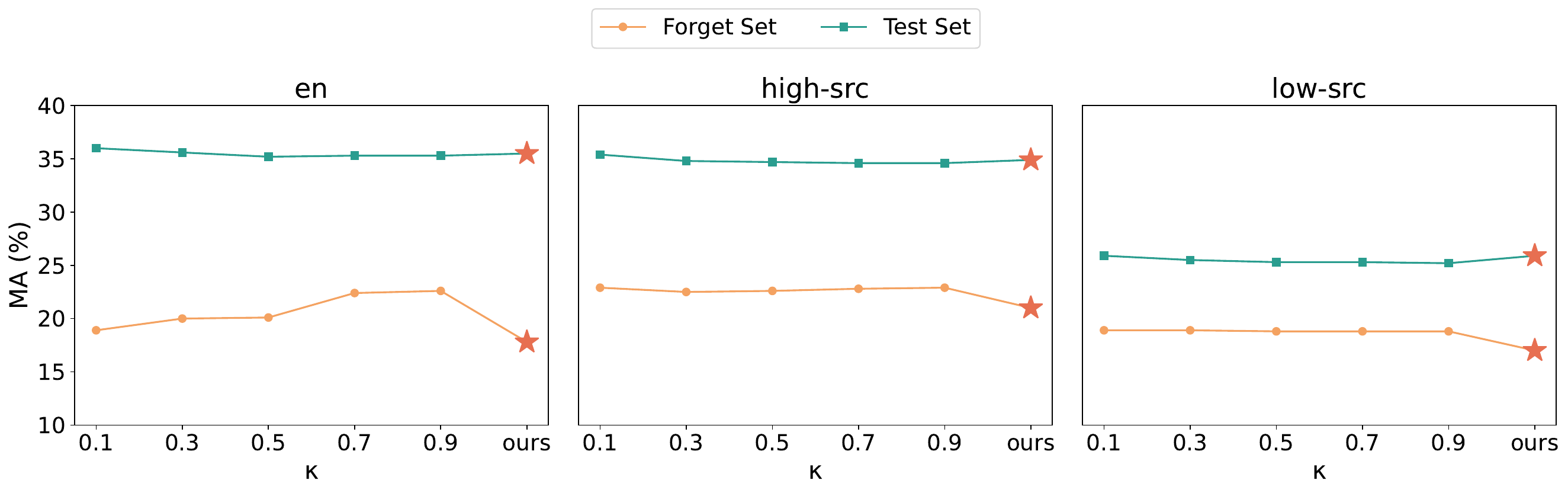}
    \caption{Comparison of the forget set and test set performance of BLOOM-3B after unlearning on FLORES-200 for \textsc{en}, \textsc{high-src}, and \textsc{low-src} across different $\kappa$ values. Our adaptive unlearning scheme yields the lowest MA on the forget set and maintains a competitive MA on the test set, highlighting the superiority of the approach.}
    \label{fig:kappa_exp}
\end{figure*}

\subsection{Effect of Adaptive Unlearning} \label{sec5.4}

To evaluate the effectiveness of our adaptive unlearning scheme, we fix various $\kappa$ values and compare them against our proposed method.
As illustrated in Figure~\ref{fig:kappa_exp}, the adaptive unlearning approach implemented in \textsc{LingTea} consistently achieves the lowest MA on the forget set across all categories, including English, high-resource, and low-resource languages.
Moreover, \textsc{LingTea} exhibits competitive performance on the test set, indicating its ability to retain knowledge effectively.
These findings demonstrate that selectively adapting to the teacher's strengths in specific languages enhances the overall multilingual unlearning process.

\subsection{Retaining World Knowledge}

\begin{figure}[h]
    \centering
    \includegraphics[width=\linewidth]{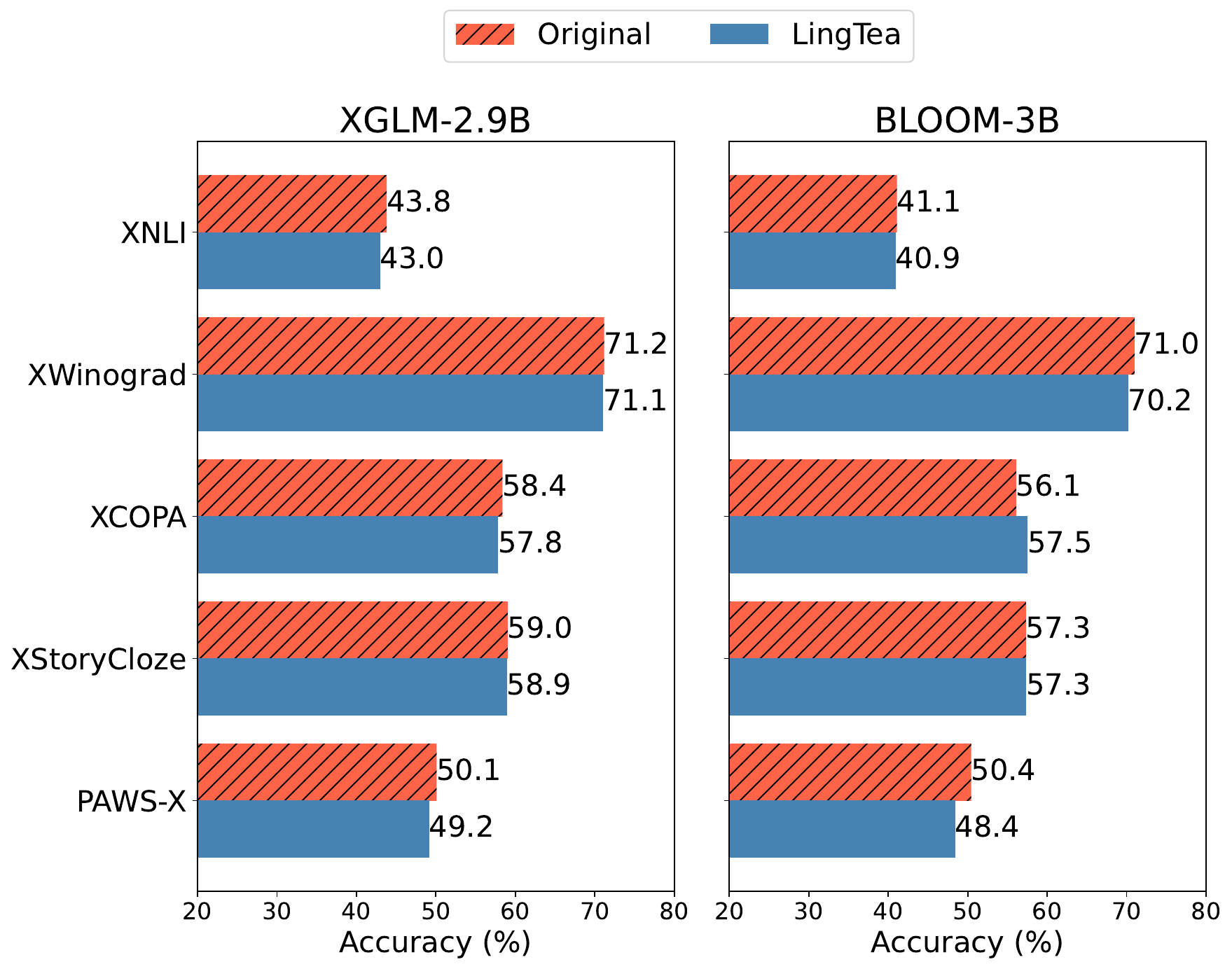}
    \caption{Zero-shot performance comparison between the original model and our \textsc{LingTea} framework across five multilingual language understanding tasks. The results demonstrate that \textsc{LingTea} retains world knowledge on par with the original model, ensuring the safety and efficacy of our unlearning approach.}
    \label{fig:world_knowledge}
\end{figure}

While our unlearning approach may succeed in retaining the test set, it is equally important to assess whether it has preserved the original multilingual language model capabilities.
To verify the retention of world knowledge, we compare our framework with the original model across five multilingual language understanding tasks: natural language inference (XNLI)~\cite{conneau2018xnli}, coreference resolution (XWinograd)~\cite{tikhonov2021xwinograd}, causal reasoning (XCOPA)~\cite{ponti2020xcopa}, sentence completion (XStoryCloze)~\cite{lin2022xglm}, and paraphrase identification (PAWS-X)~\cite{yang2019pawsx}.
We evaluate 3B models to ensure fair zero-shot performance, presenting the results in Figure~\ref{fig:world_knowledge}.
Our observations indicate that our method, \textsc{LingTea}, performs on par with the original model, thereby demonstrating the reliability of our approach.
Although NLP benchmark results may not capture all aspects of world knowledge, they at least indicate the retention of information in domains outside our unlearning data.

\subsection{Scaling the Number of Samples to Forget}

\begin{figure*}
    \centering
    \includegraphics[width=\textwidth]{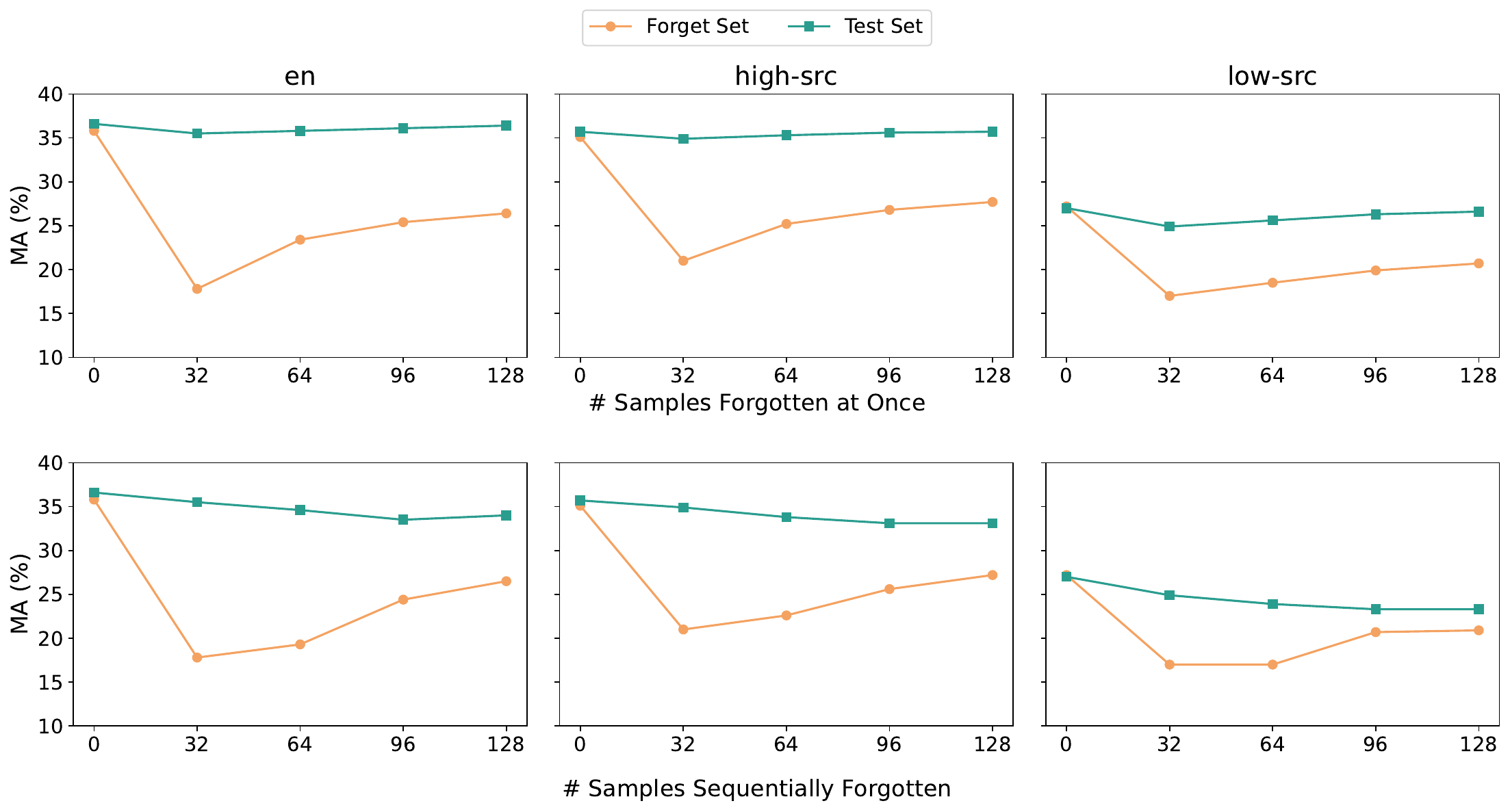}
    \caption{Performance of BLOOM-3B after unlearning token sequences in FLORES-200, shown by scaling the number of samples to be forgotten. The first row illustrates results for unlearning samples at once (\textbf{Batch Unlearning}), while the second row depicts results for unlearning samples sequentially (\textbf{Sequential Unlearning}).}
    \label{fig:scaling_exp}
\end{figure*}

To examine the scalability of our unlearning approach, we illustrate the impact of increasing the number of samples to forget by up to four-fold in Figure~\ref{fig:scaling_exp}.
Consistent with previous findings on unlearning monolingual models~\cite{jang-etal-2023-knowledge}, forgetting larger quantities of samples simultaneously proves to be more challenging, leading to no further reduction in MA.
We also investigate whether sequential unlearning could mitigate this issue; however, unlike with monolingual models, we observe no significant improvement.
On a positive note, the retention performance remains stable even as the number of samples to forget increases, highlighting the reliability of multilingual unlearning.
We hypothesize that forgetting numerous samples in a multilingual context is inherently more complex, as the total number of samples to forget effectively multiplies by the number of languages.
For instance, in the FLORES study, the increase isn't merely four-fold but rather forty-fold due to the involvement of ten languages.
Exploring the scalability of multilingual unlearning presents a non-trivial challenge, and we leave this as a direction for future research.

\section{Related Work}

\subsection{Machine Unlearning}

\citet{cao2015naivebayes} first coined the term \textit{machine unlearning}, defining it as successfully deleting an example when the outputs on a dataset are the same as if the example had never been added.
They achieved this by transforming learning algorithms into a summation form, allowing the system to forget a training data sample by updating only a few summations. 
Later, \citet{ginart2019kmeans} proposed a probabilistic definition inspired by Differential Privacy~\cite{dwork2014dp}, requiring the unlearning model to produce outputs \textit{similar} to those of a model retrained from scratch without the forgotten data.
This inspired deep approximate unlearning methods, such as those using the Fisher information matrix~\cite{golatkar2020fisher, mehta2022lcodec} and neural tangent kernel~\cite{golatkar2020ntk}.
However, these methods do not scale well, making them impractical for language models with billions of parameters.
More recently, \citet{chundawat2023badt} proposed a method using two teachers (competent and incompetent) to help a student forget certain samples while retaining the rest of the information.
\citet{kurmanji2023scrub} suggested a similar approach with a single teacher. Both methods aimed to safely forget selective samples using a teacher-student framework, primarily focusing on image classification tasks.
Our work takes a step further and considers the multilingual capabilities of the teacher, assigning language-specific weights to the distillation process.

\subsection{Knowledge Unlearning}

\citet{jang-etal-2023-knowledge} introduced \textit{knowledge unlearning}, aimed at preventing language models from generating specific token sequences. They proposed a straightforward method by inverting the original training objective of minimizing the negative log-likelihood of the token sequences.
To maintain the performance of the remaining knowledge, various works~\cite{wang-etal-2023-kga, chen2023eul, lee2024pop} employed the Kullback-Leibler (KL) divergence loss, minimizing the distributional differences between the original and unlearned models on the retained data.
Our approach builds on these methods but differs in its focus.
While the previous works targeted monolingual models like DistilBERT~\cite{sanh2019distilbert} and T5~\cite{raffel2020t5}, we extend the unlearning process to a multilingual context.

\subsection{Cross-Lingual Transfer}

Multilingual pretrained language models~\cite{devlin2019bert, conneau2020xlmr, xue2021mt5, lin2022xglm, le2023bloom} have shown to exhibit remarkable cross-lingual transfer across various tasks by leveraging shared semantic spaces and joint training techniques to bridge language gaps.
However, \citet{xu2023lime} demonstrated that editing knowledge in one language does not propagate to others and thus introduced cross-lingual model editing, a technique using random sampling of languages to improve model adaptability and robustness in a multilingual context.
Additionally, \citet{qi2023rankc} investigated the cross-lingual consistency of factual knowledge in multilingual models, finding that factual knowledge does not remain consistent across languages, but only when languages share a larger portion of vocabulary.
Building on these advancements, our work employs multilingual language models to investigate the cross-lingual transfer of machine unlearning, revealing that current unlearning methods lack this capability. 
In response, we propose an effective approach for unlearning specific information across languages, addressing the need for precise and reliable information removal in a multilingual context.

\section{Conclusion}

In response to rising privacy concerns and regulatory demands, our study pioneers a method for machine unlearning in multilingual language models. We introduce an adaptive unlearning scheme using a multilingual teacher model to address performance disparities across languages, ensuring the effective removal of sensitive information while maintaining overall model performance. Our empirical results, validated on multilingual parallel datasets, demonstrate significant improvements over existing unlearning methods. This approach not only mitigates vulnerabilities to low-resource language attacks but also offers a practical, efficient alternative to retraining models from scratch, aligning with modern privacy regulations and advancing the field of NLP.

\section*{Limitations}

Despite the robust findings presented in our paper, certain limitations warrant discussion.
The datasets used to explore multilingual unlearning in this study, namely FLORES and BMLAMA, are in the general domain.
This is due to the scarcity of multilingual parallel datasets, especially within specific domains such as privacy data.
This challenge mirrors those seen in computer vision, where datasets like CIFAR and MNIST, although unrelated to privacy, are used due to the difficulty in obtaining privacy-specific data.
Future research should focus on inventing and benchmarking real or synthetic privacy data in multilingual settings to address these gaps.
Additionally, our research was constrained by GPU resources, preventing us from testing models with 7B parameters or more. 
Investigating whether our conclusions hold for larger-scale models is a promising avenue for future work.
Lastly, our work does not handle indirect cases; that is, our framework assumes a clear data separation between forget and retain sets.
Before addressing this issue, we must consider: 1) whether we aim to remove the target tokens/sequences present in other texts, and 2) whether we should also unlearn indirect expressions.
In both cases, data augmentation could be a potential remedy in enhancing the model's robustness against indirect cases, though it may introduce unforeseen consequences.
We hope future work can address more details.

\section*{Acknowledgements}

We sincerely thank ChaeHun Park for sharing his valuable knowledge during our discussions. Additionally, we are grateful to the anonymous reviewers for their insightful feedback. This work was supported by the Institute for Information \& Communications Technology Promotion (IITP) grant funded by the Korea government (MSIT) (No.RS-2019-II190075 Artificial Intelligence Graduate School Program (KAIST)), the artificial intelligence industrial convergence cluster development project funded by the Ministry of Science and ICT (MSIT, Korea) \& Gwangju Metropolitan City, and Samsung Electronics Co., Ltd.

\bibliography{anthology,custom}
% \bibliography{custom}

\appendix

\section{Additional Details for \textsc{LingTea}} \label{sec:appendix}

% \usepackage{multirow}
% \usepackage{graphicx}
\begin{table*}[ht!]
\resizebox{\textwidth}{!}{%
\begin{tabular}{l|l|rrrrrr|rrrrrr}
\Xhline{2\arrayrulewidth}
 &
   &
  \multicolumn{6}{c|}{\textbf{Forget Set}} &
  \multicolumn{6}{c}{\textbf{Test Set}} \\
 &
   &
  \multicolumn{2}{c}{\textsc{en}} &
  \multicolumn{2}{c}{\textsc{high-src}} &
  \multicolumn{2}{c|}{\textsc{low-src}} &
  \multicolumn{2}{c}{\textsc{en}} &
  \multicolumn{2}{c}{\textsc{high-src}} &
  \multicolumn{2}{c}{\textsc{low-src}} \\ \hline
\textbf{Model} &
  \textbf{Method} &
  \multicolumn{1}{c}{\textbf{MA}($\downarrow$)} &
  \multicolumn{1}{c}{\textbf{PPL}($\uparrow$)} &
  \multicolumn{1}{c}{\textbf{MA}($\downarrow$)} &
  \multicolumn{1}{c}{\textbf{PPL}($\uparrow$)} &
  \multicolumn{1}{c}{\textbf{MA}($\downarrow$)} &
  \multicolumn{1}{c|}{\textbf{PPL}($\uparrow$)} &
  \multicolumn{1}{c}{\textbf{MA}} &
  \multicolumn{1}{c}{\textbf{PPL}} &
  \multicolumn{1}{c}{\textbf{MA}} &
  \multicolumn{1}{c}{\textbf{PPL}} &
  \multicolumn{1}{c}{\textbf{MA}} &
  \multicolumn{1}{c}{\textbf{PPL}} \\ \hline
\multirow{2}{*}{XGLM-564M} &
  GA+KL &
  23.0 &
  \textbf{7168.7} &
  24.1 &
  \textbf{2252.5} &
  21.2 &
  \textbf{2156.0} &
  30.8 &
  1192.7 &
  29.8 &
  618.8 &
  25.1 &
  853.4 \\
 &
  \textsc{LingTea} (ours) &
  \textbf{19.3} &
  4261.8 &
  \textbf{22.2} &
  816.5 &
  \textbf{19.5} &
  1031.2 &
  30.8 &
  114.8 &
  31.7 &
  85.2 &
  26.4 &
  127.8 \\ \hline
\multirow{2}{*}{XGLM-2.9B} &
  GA+KL &
  22.6 &
  1583765.3 &
  26.9 &
  \textbf{1875148.9} &
  26.5 &
  \textbf{269117.1} &
  37.2 &
  26177.9 &
  37.5 &
  12710.0 &
  33.8 &
  26767.7 \\
 &
  \textsc{LingTea} (ours) &
  \textbf{19.9} &
  \textbf{10216406.9} &
  \textbf{23.5} &
  428687.2 &
  23.1 &
  56735.2 &
  35.2 &
  102.4 &
  35.7 &
  116.4 &
  30.6 &
  172.6 \\ \hline
\multirow{2}{*}{BLOOM-560M} &
  GA+KL &
  20.2 &
  171659.1 &
  21.0 &
  \textbf{28560.2} &
  15.1 &
  \textbf{318697.5} &
  28.5 &
  26366.8 &
  27.8 &
  5924.2 &
  18.5 &
  46637.4 \\
 &
  \textsc{LingTea} (ours) &
  \textbf{18.2} &
  \textbf{2787.0} &
  \textbf{20.2} &
  1793.0 &
  \textbf{13.8} &
  6550.6 &
  28.5 &
  86.7 &
  28.6 &
  96.5 &
  19.0 &
  580.8 \\ \hline
\multirow{2}{*}{BLOOM-3B} &
  GA+KL &
  21.3 &
  359746.9 &
  21.5 &
  138652.5 &
  17.5 &
  \textbf{78245.9} &
  35.5 &
  334.8 &
  34.6 &
  206.9 &
  25.3 &
  785.1 \\
 &
  \textsc{LingTea} (ours) &
  \textbf{17.8} &
  \textbf{21063692.6} &
  \textbf{21.0} &
  \textbf{711058.2} &
  \textbf{17.0} &
  63395.3 &
  35.5 &
  51.0 &
  34.9 &
  60.0 &
  24.9 &
  233.0 \\ \Xhline{2\arrayrulewidth}
\end{tabular}%
}
\caption{Performance of unlearning multilingual token sequences on FLORES-200, comparing with GA+KL.}
\label{tab:gakl_results}
\end{table*}

\subsection{Additional Baseline Results}

We also report the GA+KL baseline, which unlearns the forget set using gradient ascent (GA) and maintains the retaining performance using KL divergence.
In the main experiment, we compare our framework with GradAscent+, which retains using gradient descent.
However, since our framework is an adaptive mixture of the two, it is equally important to compare with the pure KL baseline.
As shown in Table~\ref{tab:gakl_results}, we observe that our method \textsc{LingTea} consistently outperforms the GA+KL baseline, demonstrating the effectiveness of our approach.
This exhibits that it is very crucial to adaptively obey the teacher when unlearning in a multilingual context.

\subsection{Hyperparameters} \label{sec:hype_search}

We have performed a grid search to find the best hyperparameter configuration and report the tuning range used for our experiments in Table~\ref{tab:hyperparam_tuning_range}.
For all experiments, we have incorporated \texttt{bfloat16} mixed precision training, a linear warmup scheduler followed by decay to 0, and early stopping with a max tolerance of 5.

\begin{table}[h]
    \centering
    \begin{adjustbox}{width=\linewidth}
        \begin{tabular}{c|c|c|c}
        \toprule
            \textbf{Model} & \textbf{Hyperparameter} & \textbf{Range} & \textbf{Best}  \\
        \midrule
        \multirow{4}{*}{XGLM-564M}
            & learning rate & \{ 5e-4, 3e-4, 1e-4, 5e-5, 3e-5 \} & 5e-4  \\
            & warm-up ratio  & \{ 0.0, 0.1 \} & 0.1 \\
            & retaining samples & \{ 32, 64, 96, 128 \} & 96 \\
            & $\lambda$ & \{ 0.1, 0.5, 1.0 \} & 1.0 \\
        \midrule
        \multirow{4}{*}{XGLM-2.9B}
            & learning rate & \{ 5e-4, 3e-4, 1e-4, 5e-5, 3e-5 \} & 1e-4  \\
            & warm-up ratio  & \{ 0.0, 0.1 \} & 0.0 \\
            & retaining samples & \{ 32, 64, 96, 128 \} & 96 \\
            & $\lambda$ & \{ 0.1, 1.0, 10 \} & 1.0 \\
        \midrule
        \multirow{4}{*}{BLOOM-560M}
            & learning rate & \{ 5e-4, 3e-4, 1e-4, 5e-5, 3e-5 \} & 3e-5  \\
            & warm-up ratio  & \{ 0.0, 0.1 \} & 0.0 \\
            & retaining samples & \{ 32, 64, 96, 128 \} & 96 \\
            & $\lambda$ & \{ 0.1, 1.0, 10 \} & 1.0 \\
        \midrule
        \multirow{4}{*}{BLOOM-3B}
            & learning rate & \{ 5e-4, 3e-4, 1e-4, 5e-5, 3e-5 \} & 3e-5  \\
            & warm-up ratio  & \{ 0.0, 0.1 \} & 0.0 \\
            & retaining samples & \{ 32, 64, 96, 128 \} & 128 \\
            & $\lambda$ & \{ 0.1, 1.0, 10 \} & 1.0 \\
        \bottomrule
        \end{tabular}
    \end{adjustbox}
    \caption{Hyperparameter tuning range and best values used in the experiments.}
    \label{tab:hyperparam_tuning_range}
\end{table}

\subsection{Amount of Data Trained Per Language} \label{amount_of_langs}

The categories of high-resource, mid-resource, and low-resource languages are determined by the amount of data used to pretrain the corresponding multilingual language model.
Specifically, we follow tables in \citet{lin2022xglm} and \citet{le2023bloom} and report the statistics for the languages used in our study in Table~\ref{tab:amount_of_langs}.

\begin{table}[ht]
    \centering
    % \small
    \begin{adjustbox}{width=0.7\linewidth}
    \begin{tabular}{lcc}
    \toprule
        \textbf{Language} & XGLM & BLOOM \\
    \midrule
        English (en) & 3,324.45 & 484.95 \\
    \midrule
        \multicolumn{3}{l}{\textsc{high-src}} \\
        French (fr) & 303.76 & 208.24 \\
        Spanish (es) & 363.83 & 175.10 \\
        Chinese (zh) & 485.32 & 261.02 \\
        Portuguese (pt) & 147.12 & 79.28 \\
        Arabic (ar) & 64.34 & 74.85 \\
        Vietnamese (vi) & 50.45 & 43.71 \\
    \midrule
        \multicolumn{3}{l}{\textsc{mid-src}} \\
        Catalan (ca) & 26.90 & 17.79 \\
        Hindi (hi) & 26.63 & 24.62 \\
        Bengali (bn) & 11.19 & 18.61 \\
    \midrule
        \multicolumn{3}{l}{\textsc{low-src}} \\
        Basque (eu) & 0.35 & 2.36 \\
        Urdu (ur) & 7.77 & 2.78 \\
        Telugu (te) & 5.28 & 2.99 \\
        Swahili (sw) & 3.19 & 0.24 \\
    \bottomrule
    \end{tabular}
    \end{adjustbox}
    \caption{Amount of pretraining data in gigabytes (GB) used to train each multilingual model.}
    \label{tab:amount_of_langs}
\end{table}

\section{Per-Language Performance}

\subsection{Token Sequence Unlearning Results for Each Language} \label{flores_results_all}

We report the per-language performance of unlearning token sequences in FLORES-200 across compared models in Tables~\ref{tab:flores_forget_ma}, \ref{tab:flores_test_ma}, \ref{tab:flores_forget_ppl}, and \ref{tab:flores_test_ppl}.

% \usepackage{multirow}
\begin{table*}[]
\centering
\resizebox{0.9\textwidth}{!}{
\begin{tabular}{l|l|r|rrrrr|rrrr}
\Xhline{2\arrayrulewidth}
\textbf{} &
  \textbf{} &
  \multicolumn{1}{l|}{\textbf{}} &
  \multicolumn{5}{c|}{\textbf{High-Resource}} &
  \multicolumn{4}{c}{\textbf{Low-Resource}} \\
\textbf{Model} &
  \textbf{Method} &
  \multicolumn{1}{c|}{\textsc{en}} &
  \multicolumn{1}{c}{\textsc{fr}} &
  \multicolumn{1}{c}{\textsc{es}} &
  \multicolumn{1}{c}{\textsc{zh}} &
  \multicolumn{1}{c}{\textsc{ar}} &
  \multicolumn{1}{c|}{\textsc{vi}} &
  \multicolumn{1}{c}{\textsc{eu}} &
  \multicolumn{1}{c}{\textsc{ur}} &
  \multicolumn{1}{c}{\textsc{te}} &
  \multicolumn{1}{c}{\textsc{sw}} \\ \hline
 &
  Original &
  34.6 &
  39.2 &
  35.8 &
  32.3 &
  30.6 &
  36.2 &
  31.8 &
  30.8 &
  30.6 &
  30.0 \\
 &
  GradAscent+ &
  22.7 &
  29.3 &
  25.7 &
  \textbf{19.4} &
  21.1 &
  24.3 &
  20.9 &
  19.6 &
  20.7 &
  \textbf{18.2}\\
 &
  NegTaskVector+ &
  22.7 &
  29.3 &
  25.5 &
  21.9 &
  22.7 &
  24.5 &
  21.6 &
  22.3 &
  22.1 &
  22.1 \\
 &
  \textsc{LingTea} (ours) &
  \textbf{19.3} &
  \textbf{25.6} &
  \textbf{21.7} &
  22.3 &
  \textbf{18.8} &
  \textbf{22.6} &
  \textbf{19.7} &
  \textbf{19.5} &
  \textbf{19.9} &
  18.8 \\
\multirow{-5}{*}{XGLM-564M} &
  \cellcolor[HTML]{EFEFEF}Oracle &
  \cellcolor[HTML]{EFEFEF}17.2 &
  \cellcolor[HTML]{EFEFEF}23.7 &
  \cellcolor[HTML]{EFEFEF}19.9 &
  \cellcolor[HTML]{EFEFEF}16.7 &
  \cellcolor[HTML]{EFEFEF}14.3 &
  \cellcolor[HTML]{EFEFEF}19.2 &
  \cellcolor[HTML]{EFEFEF}14.8 &
  \cellcolor[HTML]{EFEFEF}17.4 &
  \cellcolor[HTML]{EFEFEF}17.3 &
  \cellcolor[HTML]{EFEFEF}17.1 \\ \hline
 &
  Original &
  36.8 &
  43.0 &
  36.7 &
  35.5 &
  34.1 &
  40.0 &
  38.6 &
  34.8 &
  37.6 &
  34.9 \\
 &
  GradAscent+ &
  26.3 &
  30.2 &
  29.0 &
  \textbf{22.7} &
  22.0 &
  28.2 &
  23.3 &
  \textbf{21.4} &
  \textbf{21.4} &
  23.0 \\
 &
  NegTaskVector+ &
  25.4 &
  33.2 &
  28.5 &
  25.8 &
  25.5 &
  28.6 &
  25.8 &
  25.2 &
  26.5 &
  24.7 \\
 &
  \textsc{LingTea} (ours) &
  \textbf{19.9} &
  \textbf{27.6} &
  \textbf{23.6} &
  22.8 &
  \textbf{20.6} &
  \textbf{23.1} &
  \textbf{22.6} &
  23.6 &
  24.7 &
  \textbf{21.4} \\
\multirow{-5}{*}{XGLM-2.9B} &
  \cellcolor[HTML]{EFEFEF}Oracle &
  \cellcolor[HTML]{EFEFEF}19.7 &
  \cellcolor[HTML]{EFEFEF}25.8 &
  \cellcolor[HTML]{EFEFEF}22.2 &
  \cellcolor[HTML]{EFEFEF}22.1 &
  \cellcolor[HTML]{EFEFEF}18.2 &
  \cellcolor[HTML]{EFEFEF}23.4 &
  \cellcolor[HTML]{EFEFEF}19.6 &
  \cellcolor[HTML]{EFEFEF}19.3 &
  \cellcolor[HTML]{EFEFEF}17.7 &
  \cellcolor[HTML]{EFEFEF}19.0 \\ \hline
 &
  Original &
  28.4 &
  32.3 &
  29.0 &
  21.4 &
  24.3 &
  32.2 &
  19.9 &
  23.3 &
  20.7 &
  15.9 \\
 &
  GradAscent+ &
  25.1 &
  28.4 &
  25.3 &
  16.9 &
  20.3 &
  27.5 &
  17.8 &
  19.9 &
  17.0 &
  11.4 \\
 &
  NegTaskVector+ &
  22.7 &
  25.6 &
  22.5 &
  \textbf{15.1} &
  17.9 &
  24.4 &
  14.8 &
  17.7 &
  \textbf{14.1} &
  10.2 \\
 &
  \textsc{LingTea} (ours) &
  \textbf{18.2} &
  \textbf{24.5} &
  \textbf{20.8} &
  16.2 &
  \textbf{16.9} &
  \textbf{22.6} &
  \textbf{13.1} &
  \textbf{16.8} &
  15.5 &
  \textbf{9.8} \\
\multirow{-5}{*}{BLOOM-560M} &
  \cellcolor[HTML]{EFEFEF}Oracle &
  \cellcolor[HTML]{EFEFEF}13.9 &
  \cellcolor[HTML]{EFEFEF}17.3 &
  \cellcolor[HTML]{EFEFEF}14.9 &
  \cellcolor[HTML]{EFEFEF}8.8 &
  \cellcolor[HTML]{EFEFEF}8.6 &
  \cellcolor[HTML]{EFEFEF}17.0 &
  \cellcolor[HTML]{EFEFEF}7.8 &
  \cellcolor[HTML]{EFEFEF}11.3 &
  \cellcolor[HTML]{EFEFEF}10.7 &
  \cellcolor[HTML]{EFEFEF}9.7 \\ \hline
 &
  Original &
  35.8 &
  39.6 &
  37.3 &
  30.3 &
  28.2 &
  40.1 &
  30.3 &
  28.8 &
  28.1 &
  21.5 \\
 &
  GradAscent+ &
  25.5 &
  29.3 &
  26.4 &
  18.7 &
  18.6 &
  27.3 &
  16.2 &
  \textbf{18.4} &
  \textbf{15.2} &
  \textbf{10.2} \\
 &
  NegTaskVector+ &
  28.7 &
  32.8 &
  29.6 &
  24.4 &
  20.8 &
  31.7 &
  22.2 &
  21.9 &
  20.8 &
  15.3 \\
 &
  \textsc{LingTea} (ours) &
  \textbf{17.8} &
  \textbf{23.2} &
  \textbf{22.4} &
  \textbf{18.3} &
  \textbf{17.6} &
  \textbf{23.5} &
  \textbf{15.5} &
  19.2 &
  20.3 &
  13.0 \\
\multirow{-5}{*}{BLOOM-3B} &
  \cellcolor[HTML]{EFEFEF}Oracle &
  \cellcolor[HTML]{EFEFEF}13.8 &
  \cellcolor[HTML]{EFEFEF}15.2 &
  \cellcolor[HTML]{EFEFEF}14.6 &
  \cellcolor[HTML]{EFEFEF}9.4 &
  \cellcolor[HTML]{EFEFEF}9.7 &
  \cellcolor[HTML]{EFEFEF}17.8 &
  \cellcolor[HTML]{EFEFEF}9.2 &
  \cellcolor[HTML]{EFEFEF}11.3 &
  \cellcolor[HTML]{EFEFEF}9.3 &
  \cellcolor[HTML]{EFEFEF}6.9 \\ \Xhline{2\arrayrulewidth}
\end{tabular}
}
\caption{Memorization Accuracy (\%) of Forget Set in FLORES-200.}
\label{tab:flores_forget_ma}
\end{table*}
% \usepackage{multirow}
\begin{table*}[]
\centering
\resizebox{0.9\textwidth}{!}{
\begin{tabular}{l|l|r|rrrrr|rrrr}
\Xhline{2\arrayrulewidth}
\textbf{} &
  \textbf{} &
  \multicolumn{1}{l|}{\textbf{}} &
  \multicolumn{5}{c|}{\textbf{High-Resource}} &
  \multicolumn{4}{c}{\textbf{Low-Resource}} \\
\textbf{Model} &
  \textbf{Method} &
  \multicolumn{1}{c|}{\textsc{en}} &
  \multicolumn{1}{c}{\textsc{fr}} &
  \multicolumn{1}{c}{\textsc{es}} &
  \multicolumn{1}{c}{\textsc{zh}} &
  \multicolumn{1}{c}{\textsc{ar}} &
  \multicolumn{1}{c|}{\textsc{vi}} &
  \multicolumn{1}{c}{\textsc{eu}} &
  \multicolumn{1}{c}{\textsc{ur}} &
  \multicolumn{1}{c}{\textsc{te}} &
  \multicolumn{1}{c}{\textsc{sw}} \\ \hline
 &
  Original &
  35.4 &
  40.4 &
  36.5 &
  34.0 &
  30.8 &
  34.8 &
  32.4 &
  29.5 &
  32.1 &
  28.9 \\
 &
  GradAscent+ &
  32.2 &
  37.5 &
  34.0 &
  28.4 &
  28.6 &
  31.2 &
  28.2 &
  24.8 &
  27.3 &
  24.0 \\
 &
  NegTaskVector+ &
  30.8 &
  37.6 &
  33.6 &
  30.0 &
  29.1 &
  31.1 &
  28.4 &
  27.6 &
  29.8 &
  26.3 \\
 &
  \textsc{LingTea} (ours) &
  30.8 &
  36.7 &
  32.8 &
  30.5 &
  27.5 &
  30.8 &
  27.3 &
  25.9 &
  27.4 &
  24.9 \\
\multirow{-5}{*}{XGLM-564M} &
  \cellcolor[HTML]{EFEFEF}Oracle &
  \cellcolor[HTML]{EFEFEF}32.4 &
  \cellcolor[HTML]{EFEFEF}38.1 &
  \cellcolor[HTML]{EFEFEF}34.5 &
  \cellcolor[HTML]{EFEFEF}31.8 &
  \cellcolor[HTML]{EFEFEF}29.3 &
  \cellcolor[HTML]{EFEFEF}32.5 &
  \cellcolor[HTML]{EFEFEF}30.3 &
  \cellcolor[HTML]{EFEFEF}28.6 &
  \cellcolor[HTML]{EFEFEF}29.6 &
  \cellcolor[HTML]{EFEFEF}27.7 \\ \hline
 &
  Original &
  38.5 &
  43.7 &
  39.6 &
  37.5 &
  36.2 &
  39.2 &
  38.1 &
  33.9 &
  37.1 &
  33.1 \\
 &
  GradAscent+ &
  36.2 &
  41.3 &
  37.4 &
  33.0 &
  33.2 &
  36.2 &
  34.0 &
  28.5 &
  30.9 &
  29.7 \\
 &
  NegTaskVector+ &
  33.8 &
  40.8 &
  36.6 &
  33.4 &
  34.0 &
  34.8 &
  34.3 &
  31.6 &
  34.4 &
  30.1 \\
 &
  \textsc{LingTea} (ours) &
  35.2 &
  40.7 &
  37.1 &
  34.1 &
  31.5 &
  35.0 &
  32.3 &
  29.8 &
  31.8 &
  28.6 \\
\multirow{-5}{*}{XGLM-2.9B} &
  \cellcolor[HTML]{EFEFEF}Oracle &
  \cellcolor[HTML]{EFEFEF}38.2 &
  \cellcolor[HTML]{EFEFEF}43.2 &
  \cellcolor[HTML]{EFEFEF}39.1 &
  \cellcolor[HTML]{EFEFEF}37.5 &
  \cellcolor[HTML]{EFEFEF}35.0 &
  \cellcolor[HTML]{EFEFEF}38.9 &
  \cellcolor[HTML]{EFEFEF}36.5 &
  \cellcolor[HTML]{EFEFEF}33.7 &
  \cellcolor[HTML]{EFEFEF}35.3 &
  \cellcolor[HTML]{EFEFEF}32.9 \\ \hline
 &
  Original &
  29.5 &
  33.6 &
  30.9 &
  21.6 &
  26.9 &
  31.0 &
  20.5 &
  22.6 &
  20.2 &
  14.4 \\
 &
  GradAscent+ &
  29.7 &
  33.6 &
  30.9 &
  21.4 &
  26.4 &
  30.7 &
  20.9 &
  21.9 &
  19.5 &
  14.1 \\
 &
  NegTaskVector+ &
  28.6 &
  33.0 &
  30.1 &
  20.3 &
  26.4 &
  29.7 &
  19.9 &
  22.3 &
  19.2 &
  13.9 \\
 &
  \textsc{LingTea} (ours) &
  28.5 &
  32.9 &
  30.7 &
  22.0 &
  27.0 &
  30.3 &
  20.0 &
  22.4 &
  20.0 &
  13.8 \\
\multirow{-5}{*}{BLOOM-560M} &
  \cellcolor[HTML]{EFEFEF}Oracle &
  \cellcolor[HTML]{EFEFEF}31.0 &
  \cellcolor[HTML]{EFEFEF}34.8 &
  \cellcolor[HTML]{EFEFEF}32.0 &
  \cellcolor[HTML]{EFEFEF}24.4 &
  \cellcolor[HTML]{EFEFEF}27.2 &
  \cellcolor[HTML]{EFEFEF}32.3 &
  \cellcolor[HTML]{EFEFEF}21.2 &
  \cellcolor[HTML]{EFEFEF}23.4 &
  \cellcolor[HTML]{EFEFEF}21.5 &
  \cellcolor[HTML]{EFEFEF}16.0 \\ \hline
 &
  Original &
  36.6 &
  40.7 &
  37.4 &
  29.4 &
  32.0 &
  38.8 &
  30.0 &
  28.5 &
  26.7 &
  22.8 \\
 &
  GradAscent+ &
  35.6 &
  39.7 &
  36.5 &
  27.9 &
  31.0 &
  37.3 &
  28.8 &
  26.7 &
  24.4 &
  21.1 \\
 &
  NegTaskVector+ &
  36.6 &
  40.5 &
  37.5 &
  29.4 &
  32.1 &
  38.7 &
  29.4 &
  28.7 &
  25.9 &
  22.1 \\
 &
  \textsc{LingTea} (ours) &
  35.5 &
  39.5 &
  36.6 &
  29.4 &
  31.6 &
  37.2 &
  26.9 &
  27.6 &
  25.7 &
  19.7 \\
\multirow{-5}{*}{BLOOM-3B} &
  \cellcolor[HTML]{EFEFEF}Oracle &
  \cellcolor[HTML]{EFEFEF}35.7 &
  \cellcolor[HTML]{EFEFEF}40.2 &
  \cellcolor[HTML]{EFEFEF}37.2 &
  \cellcolor[HTML]{EFEFEF}29.6 &
  \cellcolor[HTML]{EFEFEF}32.4 &
  \cellcolor[HTML]{EFEFEF}37.9 &
  \cellcolor[HTML]{EFEFEF}28.6 &
  \cellcolor[HTML]{EFEFEF}28.9 &
  \cellcolor[HTML]{EFEFEF}27.6 &
  \cellcolor[HTML]{EFEFEF}22.7 \\ \hline
\end{tabular}
}
\caption{Memorization Accuracy (\%) of Test Set in FLORES-200.}
\label{tab:flores_test_ma}
\end{table*}
% \usepackage{multirow}
\begin{table*}[]
\resizebox{\textwidth}{!}{
\begin{tabular}{l|l|r|rrrrr|rrrr}
\Xhline{2\arrayrulewidth}
\textbf{} &
  \textbf{} &
  \multicolumn{1}{l|}{\textbf{}} &
  \multicolumn{5}{c|}{\textbf{High-Resource}} &
  \multicolumn{4}{c}{\textbf{Low-Resource}} \\
\textbf{Model} &
  \textbf{Method} &
  \multicolumn{1}{c|}{\textsc{en}} &
  \multicolumn{1}{c}{\textsc{fr}} &
  \multicolumn{1}{c}{\textsc{es}} &
  \multicolumn{1}{c}{\textsc{zh}} &
  \multicolumn{1}{c}{\textsc{ar}} &
  \multicolumn{1}{c|}{\textsc{vi}} &
  \multicolumn{1}{c}{\textsc{eu}} &
  \multicolumn{1}{c}{\textsc{ur}} &
  \multicolumn{1}{c}{\textsc{te}} &
  \multicolumn{1}{c}{\textsc{sw}} \\ \hline
 &
  Original &
  117.4 &
  71.1 &
  118.1 &
  209.6 &
  151.8 &
  133.4 &
  162.8 &
  122.6 &
  85.1 &
  230.7 \\
 &
  GradAscent+ &
  4242.9 &
  \textbf{935.0} & 
  \textbf{1242.0} &
  \textbf{22843.9} &
  \textbf{2428.6} &
  \textbf{3925.0} &
  \textbf{4142.2} &
  \textbf{21813.6} &
  \textbf{8914.5} &
  \textbf{32565.1} \\
 &
  NegTaskVector+ &
  663.7 &
  253.8 &
  484.0 &
  853.6 &
  473.8 &
  544.5 &
  654.5 &
  412.6 &
  254.2 &
  872.9 \\
 &
  \textsc{LingTea} (ours) &
  \textbf{4261.8} &
  547.2 &
  1188.4 &
  882.2 &
  720.2 &
  744.8 &
  1185.5 &
  609.1 &
  843.2 &
  1487.1 \\
\multirow{-5}{*}{XGLM-564M} &
  \cellcolor[HTML]{EFEFEF}Oracle &
  \cellcolor[HTML]{EFEFEF}6295.8 &
  \cellcolor[HTML]{EFEFEF}2336.1 &
  \cellcolor[HTML]{EFEFEF}2375.9 &
  \cellcolor[HTML]{EFEFEF}9107.0 &
  \cellcolor[HTML]{EFEFEF}5546.9 &
  \cellcolor[HTML]{EFEFEF}3529.5 &
  \cellcolor[HTML]{EFEFEF}5402.9 &
  \cellcolor[HTML]{EFEFEF}4823.6 &
  \cellcolor[HTML]{EFEFEF}3879.0 &
  \cellcolor[HTML]{EFEFEF}5015.6 \\ \hline
 &
  Original &
  66.6 &
  44.9 &
  57.6 &
  231.4 &
  122.6 &
  67.3 &
  67.8 &
  120.2 &
  60.6 &
  114.0 \\
 &
  GradAscent+ &
  16553.7 &
  35430.7 &
  6453.4 &
  \textbf{17446637.7} &
  12230.1 &
  19259.8 &
  119347.7 &
  \textbf{2133682.2} &
  \textbf{4157203.2} &
  \textbf{45591.5} \\
 &
  NegTaskVector+ &
  206.8 &
  91.3 &
  157.5 &
  221.4 &
  298.0 &
  156.0 &
  266.7 &
  213.5 &
  215.9 &
  291.5 \\
 &
  \textsc{LingTea} (ours) &
  \textbf{10216406.9} &
  \textbf{89399.2} &
  \textbf{7333.6} &
  1923080.6 &
  \textbf{55752.9} &
  \textbf{67869.4} &
  \textbf{200656.9} &
  11703.3 &
  5415.2 &
  9165.5 \\
\multirow{-5}{*}{XGLM-2.9B} &
  \cellcolor[HTML]{EFEFEF}Oracle &
  \cellcolor[HTML]{EFEFEF}11605.0 &
  \cellcolor[HTML]{EFEFEF}6285.3 &
  \cellcolor[HTML]{EFEFEF}4634.5 &
  \cellcolor[HTML]{EFEFEF}171286.7 &
  \cellcolor[HTML]{EFEFEF}9716.7 &
  \cellcolor[HTML]{EFEFEF}3043.3 &
  \cellcolor[HTML]{EFEFEF}8136.9 &
  \cellcolor[HTML]{EFEFEF}682569.5 &
  \cellcolor[HTML]{EFEFEF}11299.8 &
  \cellcolor[HTML]{EFEFEF}6214.6 \\ \hline
 &
  Original &
  81.0 &
  48.5 &
  59.9 &
  151.6 &
  115.7 &
  56.3 &
  300.0 &
  183.0 &
  647.0 &
  1283.4 \\
 &
  GradAscent+ &
  127.0 &
  76.4 &
  104.6 &
  239.6 &
  189.3 &
  102.4 &
  530.7 &
  389.4 &
  2723.0 &
  4329.3 \\
 &
  NegTaskVector+ &
  277.1 &
  151.4 &
  204.4 &
  548.3 &
  369.3 &
  178.0 &
  1049.8 &
  561.3 &
  3282.7 &
  5835.6 \\
 &
  \textsc{LingTea} (ours) &
  \textbf{2787.0} &
  \textbf{1719.4} &
  \textbf{2527.2} &
  \textbf{2712.2} &
  \textbf{1072.7} &
  \textbf{933.3} &
  \textbf{6191.8} &
  \textbf{1268.3} &
  \textbf{7125.7} &
  \textbf{11616.8} \\
\multirow{-5}{*}{BLOOM-560M} &
  \cellcolor[HTML]{EFEFEF}Oracle &
  \cellcolor[HTML]{EFEFEF}12702.6 &
  \cellcolor[HTML]{EFEFEF}6706.1 &
  \cellcolor[HTML]{EFEFEF}350794.6 &
  \cellcolor[HTML]{EFEFEF}70639.8 &
  \cellcolor[HTML]{EFEFEF}31163.5 &
  \cellcolor[HTML]{EFEFEF}6724.8 &
  \cellcolor[HTML]{EFEFEF}321866.9 &
  \cellcolor[HTML]{EFEFEF}38910.3 &
  \cellcolor[HTML]{EFEFEF}45578.8 &
  \cellcolor[HTML]{EFEFEF}6366.4 \\ \hline
 &
  Original &
  42.4 &
  29.9 &
  35.1 &
  81.5 &
  82.3 &
  28.8 &
  119.5 &
  85.3 &
  123.6 &
  268.5 \\
 &
  GradAscent+ &
  291.2 &
  348.5 &
  246.9 &
  3039.6 &
  737.9 &
  191.9 &
  3872.6 &
  654.7 &
  5163.4 &
  19702.8 \\
 &
  NegTaskVector+ &
  119.7 &
  77.7 &
  90.1 &
  210.7 &
  227.3 &
  73.2 &
  496.9 &
  255.1 &
  453.9 &
  1283.0 \\
 &
  \textsc{LingTea} (ours) &
  \textbf{21063692.6} &
  \textbf{276395.7} &
  \textbf{7120.2} &
  \textbf{2312623.1} &
  \textbf{859031.7} &
  \textbf{100120.3} &
  \textbf{203023.3} &
  \textbf{16307.8} &
  \textbf{5384.3} &
  \textbf{28865.8} \\
\multirow{-5}{*}{BLOOM-3B} &
  \cellcolor[HTML]{EFEFEF}Oracle &
  \cellcolor[HTML]{EFEFEF}134342.4 &
  \cellcolor[HTML]{EFEFEF}33805.2 &
  \cellcolor[HTML]{EFEFEF}168509.4 &
  \cellcolor[HTML]{EFEFEF}1163537.3 &
  \cellcolor[HTML]{EFEFEF}219896.4 &
  \cellcolor[HTML]{EFEFEF}19421.4 &
  \cellcolor[HTML]{EFEFEF}1275923.6 &
  \cellcolor[HTML]{EFEFEF}156499.3 &
  \cellcolor[HTML]{EFEFEF}141361.8 &
  \cellcolor[HTML]{EFEFEF}297537.1 \\ \Xhline{2\arrayrulewidth}
\end{tabular}
}
\caption{Perplexity of Forget Set in FLORES-200.}
\label{tab:flores_forget_ppl}
\end{table*}
% \usepackage{multirow}
\begin{table*}[]
\resizebox{\textwidth}{!}{
\begin{tabular}{l|l|r|rrrrr|rrrr}
\Xhline{2\arrayrulewidth}
\textbf{} &
  \textbf{} &
  \multicolumn{1}{l|}{\textbf{}} &
  \multicolumn{5}{c|}{\textbf{High-Resource}} &
  \multicolumn{4}{c}{\textbf{Low-Resource}} \\
\textbf{Model} &
  \textbf{Method} &
  \multicolumn{1}{c|}{\textsc{en}} &
  \multicolumn{1}{c}{\textsc{fr}} &
  \multicolumn{1}{c}{\textsc{es}} &
  \multicolumn{1}{c}{\textsc{zh}} &
  \multicolumn{1}{c}{\textsc{ar}} &
  \multicolumn{1}{c|}{\textsc{vi}} &
  \multicolumn{1}{c}{\textsc{eu}} &
  \multicolumn{1}{c}{\textsc{ur}} &
  \multicolumn{1}{c}{\textsc{te}} &
  \multicolumn{1}{c}{\textsc{sw}} \\ \hline
 & Original       & 107.1 & 56.5  & 93.6  & 199.1  & 124.9 & 126.6 & 163.7 & 135.6 & 86.5  & 229.9  \\
 & GradAscent+    & 301.0 & 102.1 & 162.8 & 1153.5 & 409.6 & 372.2 & 563.6 & 1745.8 & 664.0 & 1705.7 \\
 & NegTaskVector+ & 191.3 & 74.5  & 125.9 & 288.8  & 160.9 & 192.3 & 226.5 & 181.7 & 111.2 & 294.4  \\
 & \textsc{LingTea} (ours) & 114.8 & 46.2  & 72.4  & 107.9  & 116.7 & 82.9  & 121.6 & 124.6 & 96.4  & 168.7  \\
\multirow{-5}{*}{XGLM-564M} &
  \cellcolor[HTML]{EFEFEF}Oracle &
  \cellcolor[HTML]{EFEFEF}114.4 &
  \cellcolor[HTML]{EFEFEF}46.8 &
  \cellcolor[HTML]{EFEFEF}65.0 &
  \cellcolor[HTML]{EFEFEF}121.2 &
  \cellcolor[HTML]{EFEFEF}120.4 &
  \cellcolor[HTML]{EFEFEF}81.1 &
  \cellcolor[HTML]{EFEFEF}113.2 &
  \cellcolor[HTML]{EFEFEF}104.6 &
  \cellcolor[HTML]{EFEFEF}93.8 &
  \cellcolor[HTML]{EFEFEF}141.3 \\ \hline
 & Original       & 68.6  & 35.7  & 48.7  & 192.3  & 92.6  & 81.5  & 71.6  & 131.5 & 64.3  & 128.8  \\
 & GradAscent+    & 922.8 & 184.5 & 258.7 & 2573.7 & 403.9 & 531.5 & 873.0 & 15338.7 & 10443.3 & 1081.6 \\
 & NegTaskVector+ & 78.4  & 30.7  & 52.6  & 79.3   & 76.2  & 60.3  & 80.9  & 89.3  & 81.5  & 112.6  \\
 & \textsc{LingTea} (ours) & 102.4 & 40.3  & 59.0  & 171.9  & 221.3 & 89.6  & 111.5 & 254.0 & 155.5 & 169.5  \\
\multirow{-5}{*}{XGLM-2.9B} &
  \cellcolor[HTML]{EFEFEF}Oracle &
  \cellcolor[HTML]{EFEFEF}70.5 &
  \cellcolor[HTML]{EFEFEF}32.7 &
  \cellcolor[HTML]{EFEFEF}47.4 &
  \cellcolor[HTML]{EFEFEF}62.5 &
  \cellcolor[HTML]{EFEFEF}73.9 &
  \cellcolor[HTML]{EFEFEF}40.1 &
  \cellcolor[HTML]{EFEFEF}70.2 &
  \cellcolor[HTML]{EFEFEF}65.5 &
  \cellcolor[HTML]{EFEFEF}60.9 &
  \cellcolor[HTML]{EFEFEF}88.2 \\ \hline
 & Original       & 73.2  & 39.8  & 48.3  & 154.6  & 97.1  & 52.2  & 311.4 & 178.8 & 558.7 & 1213.8 \\
 & GradAscent+    & 72.4  & 39.5  & 48.3  & 158.3  & 103.0 & 55.3  & 319.6 & 212.1 & 826.5 & 1385.6 \\
 & NegTaskVector+ & 83.0  & 43.9  & 54.3  & 170.3  & 114.8 & 64.6  & 350.9 & 217.7 & 806.3 & 1518.8 \\
 & \textsc{LingTea} (ours) & 86.7  & 47.9  & 59.9  & 199.9  & 106.7 & 67.8  & 376.5 & 225.9 & 559.9 & 1160.7 \\
\multirow{-5}{*}{BLOOM-560M} &
  \cellcolor[HTML]{EFEFEF}Oracle &
  \cellcolor[HTML]{EFEFEF}71.8 &
  \cellcolor[HTML]{EFEFEF}39.1 &
  \cellcolor[HTML]{EFEFEF}62.5 &
  \cellcolor[HTML]{EFEFEF}164.7 &
  \cellcolor[HTML]{EFEFEF}108.1 &
  \cellcolor[HTML]{EFEFEF}57.6 &
  \cellcolor[HTML]{EFEFEF}390.4 &
  \cellcolor[HTML]{EFEFEF}217.2 &
  \cellcolor[HTML]{EFEFEF}468.8 &
  \cellcolor[HTML]{EFEFEF}664.5 \\ \hline
 & Original       & 42.7  & 24.6  & 30.1  & 83.2   & 60.2  & 28.8  & 114.8 & 89.9  & 119.5 & 295.4  \\
 & GradAscent+    & 54.5  & 28.5  & 33.1  & 143.5  & 86.5  & 38.1  & 179.3 & 155.3 & 309.9 & 602.7  \\
 & NegTaskVector+ & 42.8  & 24.7  & 30.5  & 78.4   & 60.6  & 28.7  & 122.6 & 93.2  & 128.6 & 330.7  \\
 & \textsc{LingTea} (ours) & 51.0  & 27.5  & 33.6  & 114.8  & 83.9  & 40.3  & 153.9 & 127.1 & 184.7 & 466.5  \\
\multirow{-5}{*}{BLOOM-3B} &
  \cellcolor[HTML]{EFEFEF}Oracle &
  \cellcolor[HTML]{EFEFEF}49.5 &
  \cellcolor[HTML]{EFEFEF}25.7 &
  \cellcolor[HTML]{EFEFEF}32.2 &
  \cellcolor[HTML]{EFEFEF}99.7 &
  \cellcolor[HTML]{EFEFEF}68.0 &
  \cellcolor[HTML]{EFEFEF}33.6 &
  \cellcolor[HTML]{EFEFEF}129.5 &
  \cellcolor[HTML]{EFEFEF}108.2 &
  \cellcolor[HTML]{EFEFEF}141.8 &
  \cellcolor[HTML]{EFEFEF}268.3 \\ \Xhline{2\arrayrulewidth}
\end{tabular}
}
\caption{Perplexity of Test Set in FLORES-200.}
\label{tab:flores_test_ppl}
\end{table*}

\subsection{Factual Knowledge Unlearning Results for Each Language} \label{bmlama_results_all}

We report the per-language performance of unlearning factual knowledge in BMLAMA-53 across compared models in Tables~\ref{tab:bmlama_forget_pa}, \ref{tab:bmlama_test_pa}, \ref{tab:bmlama_forget_ppl}, and \ref{tab:bmlama_test_ppl}.

% \usepackage{multirow}
\begin{table*}[]
\centering
\resizebox{0.9\textwidth}{!}{
\begin{tabular}{l|l|r|rrrrr|rrr}
\Xhline{2\arrayrulewidth}
\textbf{} &
  \textbf{} &
  \multicolumn{1}{l|}{\textbf{}} &
  \multicolumn{5}{c|}{\textbf{High-Resource}} &
  \multicolumn{3}{c}{\textbf{Mid-Resource}} \\
\textbf{Model} &
  \textbf{Method} &
  \multicolumn{1}{c|}{\textsc{en}} &
  \multicolumn{1}{c}{\textsc{fr}} &
  \multicolumn{1}{c}{\textsc{es}} &
  \multicolumn{1}{c}{\textsc{pt}} &
  \multicolumn{1}{c}{\textsc{ar}} &
  \multicolumn{1}{c|}{\textsc{vi}} &
  \multicolumn{1}{c}{\textsc{ca}} &
  \multicolumn{1}{c}{\textsc{hi}} &
  \multicolumn{1}{c}{\textsc{bn}} \\ \hline
 &
  Original &
  28.1 &
  12.5 &
  12.5 &
  12.5 &
  12.5 &
  18.8 &
  15.6 &
  21.9 &
  18.8 \\
 &
  GradAscent+ &
  27.1 &
  3.1 &
  \textbf{3.1} &
  \textbf{6.3} &
  6.3 &
  17.7 &
  10.4 &
  15.6 &
  \textbf{11.5} \\
 &
  NegTaskVector+ &
  28.1 &
  3.1 &
  6.3 &
  \textbf{6.3} &
  \textbf{3.1} &
  17.7 &
  10.4 &
  \textbf{11.5} &
  13.5 \\
 &
  \textsc{LingTea} (ours) &
  \textbf{25.0} &
  \textbf{0.0} &
  \textbf{3.1} &
  \textbf{6.3} &
  4.2 &
  \textbf{13.5} &
  \textbf{4.2} &
  15.6 &
  12.5 \\
\multirow{-5}{*}{XGLM-564M} &
  \cellcolor[HTML]{EFEFEF}Oracle &
  \cellcolor[HTML]{EFEFEF}5.2 &
  \cellcolor[HTML]{EFEFEF}0.0 &
  \cellcolor[HTML]{EFEFEF}3.1 &
  \cellcolor[HTML]{EFEFEF}6.3 &
  \cellcolor[HTML]{EFEFEF}0.0 &
  \cellcolor[HTML]{EFEFEF}3.1 &
  \cellcolor[HTML]{EFEFEF}1.0 &
  \cellcolor[HTML]{EFEFEF}6.3 &
  \cellcolor[HTML]{EFEFEF}0.0 \\ \hline
 &
  Original &
  34.4 &
  6.3 &
  12.5 &
  15.6 &
  15.6 &
  28.1 &
  25.0 &
  31.3 &
  18.8 \\
 &
  GradAscent+ &
  29.2 &
  7.3 &
  8.3 &
  10.4 &
  7.3 &
  21.9 &
  12.5 &
  15.6 &
  \textbf{7.3} \\
 &
  NegTaskVector+ &
  29.2 &
  7.3 &
  6.3 &
  8.3 &
  8.3 &
  16.7 &
  \textbf{8.3} &
  \textbf{17.7} &
  11.5 \\
 &
  \textsc{LingTea} (ours) &
  \textbf{14.6} &
  \textbf{4.2} &
  \textbf{4.2} &
  \textbf{6.3} &
  \textbf{6.3} &
  \textbf{13.5} &
  13.5 &
  12.5 &
  12.5 \\
\multirow{-5}{*}{XGLM-2.9B} &
  \cellcolor[HTML]{EFEFEF}Oracle &
  \cellcolor[HTML]{EFEFEF}13.5 &
  \cellcolor[HTML]{EFEFEF}4.2 &
  \cellcolor[HTML]{EFEFEF}5.2 &
  \cellcolor[HTML]{EFEFEF}10.4 &
  \cellcolor[HTML]{EFEFEF}6.3 &
  \cellcolor[HTML]{EFEFEF}1.0 &
  \cellcolor[HTML]{EFEFEF}2.1 &
  \cellcolor[HTML]{EFEFEF}4.2 &
  \cellcolor[HTML]{EFEFEF}7.3 \\ \hline
 &
  Original &
  31.3 &
  18.8 &
  21.9 &
  18.8 &
  6.3 &
  28.1 &
  9.4 &
  12.5 &
  9.4 \\
 &
  GradAscent+ &
  15.6 &
  12.5 &
  11.5 &
  7.3 &
  \textbf{5.2} &
  19.8 &
  \textbf{4.2} &
  7.3 &
  9.4 \\
 &
  NegTaskVector+ &
  22.9 &
  15.6 &
  9.4 &
  11.5 &
  \textbf{5.2} &
  22.9 &
  5.2 &
  7.3 &
  9.4 \\
 &
  \textsc{LingTea} (ours) &
  \textbf{9.4} &
  \textbf{8.3} &
  \textbf{8.3} &
  \textbf{4.2} &
  \textbf{5.2} &
  \textbf{8.3} &
  5.2 &
  \textbf{6.3} &
  \textbf{5.2} \\
\multirow{-5}{*}{BLOOM-560M} &
  \cellcolor[HTML]{EFEFEF}Oracle &
  \cellcolor[HTML]{EFEFEF}7.3 &
  \cellcolor[HTML]{EFEFEF}3.1 &
  \cellcolor[HTML]{EFEFEF}3.1 &
  \cellcolor[HTML]{EFEFEF}4.2 &
  \cellcolor[HTML]{EFEFEF}0.0 &
  \cellcolor[HTML]{EFEFEF}3.1 &
  \cellcolor[HTML]{EFEFEF}3.1 &
  \cellcolor[HTML]{EFEFEF}0.0 &
  \cellcolor[HTML]{EFEFEF}0.0 \\ \hline
 &
  Original &
  50.0 &
  28.1 &
  28.1 &
  18.8 &
  15.6 &
  31.3 &
  18.8 &
  9.4 &
  15.6 \\
 &
  GradAscent+ &
  \textbf{16.7} &
  10.4 &
  8.3 &
  6.3 &
  \textbf{3.1} &
  10.4 &
  \textbf{3.1} &
  6.3 &
  8.3 \\
 &
  NegTaskVector+ &
  35.4 &
  16.7 &
  19.8 &
  12.5 &
  7.3 &
  24.0 &
  5.2 &
  \textbf{5.2} &
  11.5 \\
 &
  \textsc{LingTea} (ours) &
  19.8 &
  \textbf{6.3} &
  \textbf{6.3} &
  \textbf{4.2} &
  7.3 &
  \textbf{7.3} &
  5.2 &
  9.4 &
  \textbf{6.3} \\
\multirow{-5}{*}{BLOOM-3B} &
  \cellcolor[HTML]{EFEFEF}Oracle &
  \cellcolor[HTML]{EFEFEF}17.7 &
  \cellcolor[HTML]{EFEFEF}6.3 &
  \cellcolor[HTML]{EFEFEF}12.5 &
  \cellcolor[HTML]{EFEFEF}8.3 &
  \cellcolor[HTML]{EFEFEF}1.0 &
  \cellcolor[HTML]{EFEFEF}8.3 &
  \cellcolor[HTML]{EFEFEF}4.2 &
  \cellcolor[HTML]{EFEFEF}3.1 &
  \cellcolor[HTML]{EFEFEF}0.0 \\ \Xhline{2\arrayrulewidth}
\end{tabular}
}
\caption{Probing Accuracy (\%) of Forget Set in BMLAMA-53.}
\label{tab:bmlama_forget_pa}
\end{table*}
% \usepackage{multirow}
\begin{table*}[]
\centering
\resizebox{0.9\textwidth}{!}{
\begin{tabular}{l|l|r|rrrrr|rrr}
\Xhline{2\arrayrulewidth}
\textbf{} &
  \textbf{} &
  \multicolumn{1}{l|}{\textbf{}} &
  \multicolumn{5}{c|}{\textbf{High-Resource}} &
  \multicolumn{3}{c}{\textbf{Mid-Resource}} \\
\textbf{Model} &
  \textbf{Method} &
  \multicolumn{1}{c|}{\textsc{en}} &
  \multicolumn{1}{c}{\textsc{fr}} &
  \multicolumn{1}{c}{\textsc{es}} &
  \multicolumn{1}{c}{\textsc{pt}} &
  \multicolumn{1}{c}{\textsc{ar}} &
  \multicolumn{1}{c|}{\textsc{vi}} &
  \multicolumn{1}{c}{\textsc{ca}} &
  \multicolumn{1}{c}{\textsc{hi}} &
  \multicolumn{1}{c}{\textsc{bn}} \\ \hline
 &
  Original &
  29.9 &
  15.9 &
  17.1 &
  16.8 &
  16.1 &
  18.9 &
  21.4 &
  15.9 &
  15.3 \\
 &
  GradAscent+ &
  30.3 &
  15.7 &
  17.3 &
  15.3 &
  15.8 &
  19.3 &
  20.5 &
  14.3 &
  14.9 \\
 &
  NegTaskVector+ &
  30.1 &
  16.9 &
  18.0 &
  18.0 &
  16.5 &
  20.6 &
  21.0 &
  15.3 &
  16.2 \\
 &
  \textsc{LingTea} (ours) &
  29.4 &
  17.5 &
  18.1 &
  16.9 &
  15.9 &
  18.2 &
  20.2 &
  13.8 &
  16.6 \\
\multirow{-5}{*}{XGLM-564M} &
  \cellcolor[HTML]{EFEFEF}Oracle &
  \cellcolor[HTML]{EFEFEF}28.6 &
  \cellcolor[HTML]{EFEFEF}14.0 &
  \cellcolor[HTML]{EFEFEF}16.1 &
  \cellcolor[HTML]{EFEFEF}14.9 &
  \cellcolor[HTML]{EFEFEF}13.9 &
  \cellcolor[HTML]{EFEFEF}22.6 &
  \cellcolor[HTML]{EFEFEF}19.7 &
  \cellcolor[HTML]{EFEFEF}12.0 &
  \cellcolor[HTML]{EFEFEF}13.4 \\ \hline
 &
  Original &
  34.7 &
  19.3 &
  24.2 &
  25.5 &
  18.1 &
  22.6 &
  27.1 &
  15.6 &
  15.8 \\
 &
  GradAscent+ &
  35.4 &
  21.1 &
  23.0 &
  23.4 &
  16.4 &
  22.4 &
  26.3 &
  14.0 &
  12.8 \\
 &
  NegTaskVector+ &
  33.4 &
  19.0 &
  21.6 &
  21.1 &
  18.5 &
  21.8 &
  24.6 &
  14.8 &
  17.1 \\
 &
  \textsc{LingTea} (ours) &
  37.1 &
  25.2 &
  29.6 &
  25.4 &
  17.8 &
  24.3 &
  29.4 &
  17.5 &
  17.8 \\
\multirow{-5}{*}{XGLM-2.9B} &
  \cellcolor[HTML]{EFEFEF}Oracle &
  \cellcolor[HTML]{EFEFEF}43.3 &
  \cellcolor[HTML]{EFEFEF}23.4 &
  \cellcolor[HTML]{EFEFEF}28.3 &
  \cellcolor[HTML]{EFEFEF}30.6 &
  \cellcolor[HTML]{EFEFEF}20.1 &
  \cellcolor[HTML]{EFEFEF}32.8 &
  \cellcolor[HTML]{EFEFEF}35.9 &
  \cellcolor[HTML]{EFEFEF}18.7 &
  \cellcolor[HTML]{EFEFEF}17.6 \\ \hline
 &
  Original &
  28.5 &
  16.6 &
  21.0 &
  17.7 &
  11.2 &
  20.2 &
  16.0 &
  11.2 &
  9.9 \\
 &
  GradAscent+ &
  28.5 &
  15.3 &
  18.9 &
  17.3 &
  11.2 &
  20.9 &
  15.3 &
  10.1 &
  9.6 \\
 &
  NegTaskVector+ &
  29.0 &
  16.8 &
  20.4 &
  16.6 &
  11.3 &
  21.5 &
  16.0 &
  10.5 &
  9.8 \\
 &
  \textsc{LingTea} (ours) &
  27.4 &
  16.7 &
  19.3 &
  17.4 &
  11.2 &
  20.2 &
  16.2 &
  10.5 &
  9.8 \\
\multirow{-5}{*}{BLOOM-560M} &
  \cellcolor[HTML]{EFEFEF}Oracle &
  \cellcolor[HTML]{EFEFEF}29.6 &
  \cellcolor[HTML]{EFEFEF}18.5 &
  \cellcolor[HTML]{EFEFEF}19.3 &
  \cellcolor[HTML]{EFEFEF}18.1 &
  \cellcolor[HTML]{EFEFEF}10.9 &
  \cellcolor[HTML]{EFEFEF}23.5 &
  \cellcolor[HTML]{EFEFEF}15.2 &
  \cellcolor[HTML]{EFEFEF}9.8 &
  \cellcolor[HTML]{EFEFEF}10.1 \\ \hline
 &
  Original &
  46.6 &
  26.4 &
  31.3 &
  29.5 &
  16.8 &
  30.2 &
  24.1 &
  13.3 &
  10.8 \\
 &
  GradAscent+ &
  40.8 &
  21.9 &
  25.2 &
  25.5 &
  15.2 &
  30.1 &
  21.4 &
  12.1 &
  11.1 \\
 &
  NegTaskVector+ &
  47.2 &
  23.0 &
  30.0 &
  25.9 &
  15.0 &
  29.5 &
  21.7 &
  11.4 &
  10.6 \\
 &
  \textsc{LingTea} (ours) &
  47.1 &
  34.2 &
  36.1 &
  33.8 &
  19.8 &
  36.0 &
  30.5 &
  13.5 &
  10.9 \\
\multirow{-5}{*}{BLOOM-3B} &
  \cellcolor[HTML]{EFEFEF}Oracle &
  \cellcolor[HTML]{EFEFEF}46.1 &
  \cellcolor[HTML]{EFEFEF}42.2 &
  \cellcolor[HTML]{EFEFEF}39.6 &
  \cellcolor[HTML]{EFEFEF}33.7 &
  \cellcolor[HTML]{EFEFEF}19.8 &
  \cellcolor[HTML]{EFEFEF}39.0 &
  \cellcolor[HTML]{EFEFEF}32.2 &
  \cellcolor[HTML]{EFEFEF}16.3 &
  \cellcolor[HTML]{EFEFEF}12.0 \\ \Xhline{2\arrayrulewidth}
\end{tabular}
}
\caption{Probing Accuracy (\%) of Test Set in BMLAMA-53.}
\label{tab:bmlama_test_pa}
\end{table*}
% \usepackage{multirow}
\begin{table*}[]
\resizebox{\textwidth}{!}{
\begin{tabular}{l|l|r|rrrrr|rrr}
\Xhline{2\arrayrulewidth}
\textbf{} &
  \textbf{} &
  \multicolumn{1}{l|}{\textbf{}} &
  \multicolumn{5}{c|}{\textbf{High-Resource}} &
  \multicolumn{3}{c}{\textbf{Mid-Resource}} \\
\textbf{Model} &
  \textbf{Method} &
  \multicolumn{1}{c|}{\textsc{en}} &
  \multicolumn{1}{c}{\textsc{fr}} &
  \multicolumn{1}{c}{\textsc{es}} &
  \multicolumn{1}{c}{\textsc{pt}} &
  \multicolumn{1}{c}{\textsc{ar}} &
  \multicolumn{1}{c|}{\textsc{vi}} &
  \multicolumn{1}{c}{\textsc{ca}} &
  \multicolumn{1}{c}{\textsc{hi}} &
  \multicolumn{1}{c}{\textsc{bn}} \\ \hline
 &
  Original &
  122.0 &
  108.7 &
  151.4 &
  100.6 &
  110.4 &
  113.1 &
  114.7 &
  73.5 &
  48.0 \\
 &
  GradAscent+ &
  \textbf{187.2} &
  173.2 &
  262.3 &
  156.6 &
  118.1 &
  159.5 &
  170.1 &
  77.8 &
  51.3 \\
 &
  NegTaskVector+ &
  150.7 &
  136.3 &
  194.0 &
  116.4 &
  \textbf{131.9} &
  132.0 &
  122.2 &
  \textbf{90.6} &
  \textbf{59.2} \\
 &
  \textsc{LingTea} (ours) &
  185.3 &
  \textbf{177.6} &
  \textbf{282.1} &
  \textbf{168.0} &
  103.2 &
  \textbf{167.1} &
  \textbf{176.4} &
  72.7 &
  57.1 \\
\multirow{-5}{*}{XGLM-564M} &
  \cellcolor[HTML]{EFEFEF}Oracle &
  \cellcolor[HTML]{EFEFEF}71681.6 &
  \cellcolor[HTML]{EFEFEF}192.6 &
  \cellcolor[HTML]{EFEFEF}196.4 &
  \cellcolor[HTML]{EFEFEF}137.0 &
  \cellcolor[HTML]{EFEFEF}195.8 &
  \cellcolor[HTML]{EFEFEF}15205.2 &
  \cellcolor[HTML]{EFEFEF}1241.5 &
  \cellcolor[HTML]{EFEFEF}568.3 &
  \cellcolor[HTML]{EFEFEF}405.8 \\ \hline
 &
  Original &
  90.9 &
  80.7 &
  109.4 &
  84.7 &
  45.3 &
  93.6 &
  77.6 &
  36.4 &
  31.7 \\
 &
  GradAscent+ &
  133.5 &
  125.8 &
  244.5 &
  166.2 &
  \textbf{352.6} &
  140.5 &
  126.6 &
  195.5 &
  244.3 \\
 &
  NegTaskVector+ &
  124.6 &
  121.4 &
  179.3 &
  140.8 &
  60.9 &
  133.9 &
  118.8 &
  54.0 &
  43.9 \\
 &
  \textsc{LingTea} (ours) &
  \textbf{908.6} &
  \textbf{744.5} &
  \textbf{673.2} &
  \textbf{766.7} &
  163.9 &
  \textbf{1043.4} &
  \textbf{483.2} &
  \textbf{356.8} &
  \textbf{600.0} \\
\multirow{-5}{*}{XGLM-2.9B} &
  \cellcolor[HTML]{EFEFEF}Oracle &
  \cellcolor[HTML]{EFEFEF}1274.7 &
  \cellcolor[HTML]{EFEFEF}100.9 &
  \cellcolor[HTML]{EFEFEF}250.8 &
  \cellcolor[HTML]{EFEFEF}290.4 &
  \cellcolor[HTML]{EFEFEF}85.9 &
  \cellcolor[HTML]{EFEFEF}2035.9 &
  \cellcolor[HTML]{EFEFEF}1228.4 &
  \cellcolor[HTML]{EFEFEF}6126.5 &
  \cellcolor[HTML]{EFEFEF}1593.2 \\ \hline
 &
  Original &
  145.8 &
  112.4 &
  170.3 &
  227.0 &
  85.7 &
  129.4 &
  239.6 &
  183.3 &
  379.6 \\
 &
  GradAscent+ &
  238.8 &
  \textbf{174.5} &
  293.0 &
  369.0 &
  98.5 &
  167.7 &
  403.7 &
  225.9 &
  464.2 \\
 &
  NegTaskVector+ &
  184.9 &
  128.0 &
  211.3 &
  272.2 &
  91.0 &
  137.8 &
  289.7 &
  236.3 &
  468.2 \\
 &
  \textsc{LingTea} (ours) &
  \textbf{267.5} &
  174.2 &
  \textbf{354.9} &
  \textbf{486.0} &
  \textbf{113.4} &
  \textbf{209.9} &
  \textbf{471.3} &
  \textbf{313.8} &
  \textbf{691.1} \\
\multirow{-5}{*}{BLOOM-560M} &
  \cellcolor[HTML]{EFEFEF}Oracle &
  \cellcolor[HTML]{EFEFEF}629.3 &
  \cellcolor[HTML]{EFEFEF}159.6 &
  \cellcolor[HTML]{EFEFEF}316.1 &
  \cellcolor[HTML]{EFEFEF}536.6 &
  \cellcolor[HTML]{EFEFEF}168.9 &
  \cellcolor[HTML]{EFEFEF}32890.0 &
  \cellcolor[HTML]{EFEFEF}257.1 &
  \cellcolor[HTML]{EFEFEF}506.5 &
  \cellcolor[HTML]{EFEFEF}1704.6 \\ \hline
 &
  Original &
  68.9 &
  71.1 &
  80.6 &
  115.9 &
  45.8 &
  60.8 &
  99.2 &
  77.5 &
  108.3 \\
 &
  GradAscent+ &
  645.1 &
  515.1 &
  1045.2 &
  1090.3 &
  157.3 &
  280.6 &
  \textbf{682.4} &
  229.9 &
  294.9 \\
 &
  NegTaskVector+ &
  110.8 &
  118.3 &
  150.4 &
  221.1 &
  64.5 &
  87.9 &
  179.1 &
  168.3 &
  202.8 \\
 &
  \textsc{LingTea} (ours) &
  \textbf{1077.3} &
  \textbf{585.5} &
  \textbf{1243.3} &
  \textbf{1606.7} &
  \textbf{185.9} &
  \textbf{287.7} &
  621.1 &
  \textbf{400.6} &
  \textbf{1156.0} \\
\multirow{-5}{*}{BLOOM-3B} &
  \cellcolor[HTML]{EFEFEF}Oracle &
  \cellcolor[HTML]{EFEFEF}2708.9 &
  \cellcolor[HTML]{EFEFEF}398.0 &
  \cellcolor[HTML]{EFEFEF}555.1 &
  \cellcolor[HTML]{EFEFEF}506.2 &
  \cellcolor[HTML]{EFEFEF}61.5 &
  \cellcolor[HTML]{EFEFEF}405.0 &
  \cellcolor[HTML]{EFEFEF}304.1 &
  \cellcolor[HTML]{EFEFEF}3234.7 &
  \cellcolor[HTML]{EFEFEF}1795.7 \\ \Xhline{2\arrayrulewidth}
\end{tabular}
}
\caption{Perplexity of Forget Set in BMLAMA-53.}
\label{tab:bmlama_forget_ppl}
\end{table*}
% \usepackage{multirow}
\begin{table*}[]
\resizebox{\textwidth}{!}{
\begin{tabular}{l|l|r|rrrrr|rrr}
\Xhline{2\arrayrulewidth}
\textbf{} &
  \textbf{} &
  \multicolumn{1}{l|}{\textbf{}} &
  \multicolumn{5}{c|}{\textbf{High-Resource}} &
  \multicolumn{3}{c}{\textbf{Mid-Resource}} \\
\textbf{Model} &
  \textbf{Method} &
  \multicolumn{1}{c|}{\textsc{en}} &
  \multicolumn{1}{c}{\textsc{fr}} &
  \multicolumn{1}{c}{\textsc{es}} &
  \multicolumn{1}{c}{\textsc{pt}} &
  \multicolumn{1}{c}{\textsc{ar}} &
  \multicolumn{1}{c|}{\textsc{vi}} &
  \multicolumn{1}{c}{\textsc{ca}} &
  \multicolumn{1}{c}{\textsc{hi}} &
  \multicolumn{1}{c}{\textsc{bn}} \\ \hline
 & Original       & 152.8 & 122.2 & 170.4 & 113.9 & 115.2 & 153.2 & 122.5 & 99.3  & 66.0  \\
 & GradAscent+    & 187.8 & 154.3 & 223.5 & 140.6 & 110.8 & 184.8 & 149.0 & 91.4  & 62.0  \\
 & NegTaskVector+ & 145.9 & 120.4 & 165.0 & 102.5 & 116.4 & 145.6 & 105.7 & 99.0  & 66.3  \\
 & \textsc{LingTea} (ours) & 165.5 & 139.4 & 196.3 & 112.6 & 92.6  & 153.2 & 131.5 & 73.8  & 60.7  \\
\multirow{-5}{*}{XGLM-564M} &
  \cellcolor[HTML]{EFEFEF}Oracle &
  \cellcolor[HTML]{EFEFEF}1249.1 &
  \cellcolor[HTML]{EFEFEF}141.7 &
  \cellcolor[HTML]{EFEFEF}185.1 &
  \cellcolor[HTML]{EFEFEF}135.9 &
  \cellcolor[HTML]{EFEFEF}148.6 &
  \cellcolor[HTML]{EFEFEF}1226.1 &
  \cellcolor[HTML]{EFEFEF}224.1 &
  \cellcolor[HTML]{EFEFEF}222.0 &
  \cellcolor[HTML]{EFEFEF}150.3 \\ \hline
 & Original       & 112.7 & 95.1  & 121.5 & 94.1  & 51.6  & 114.2 & 85.7  & 52.9  & 38.8  \\
 & GradAscent+    & 127.5 & 116.1 & 170.1 & 125.8 & 314.0 & 146.4 & 108.5 & 204.4 & 156.0 \\
 & NegTaskVector+ & 120.2 & 110.1 & 142.7 & 109.7 & 57.3  & 128.8 & 95.6  & 58.9  & 39.9  \\
 & \textsc{LingTea} (ours) & 156.5 & 104.5 & 129.0 & 128.3 & 73.7  & 251.9 & 94.8  & 135.0 & 164.4 \\
\multirow{-5}{*}{XGLM-2.9B} &
  \cellcolor[HTML]{EFEFEF}Oracle &
  \cellcolor[HTML]{EFEFEF}176.1 &
  \cellcolor[HTML]{EFEFEF}83.3 &
  \cellcolor[HTML]{EFEFEF}92.2 &
  \cellcolor[HTML]{EFEFEF}67.4 &
  \cellcolor[HTML]{EFEFEF}79.5 &
  \cellcolor[HTML]{EFEFEF}216.8 &
  \cellcolor[HTML]{EFEFEF}80.9 &
  \cellcolor[HTML]{EFEFEF}235.6 &
  \cellcolor[HTML]{EFEFEF}84.3 \\ \hline
 & Original       & 202.6 & 134.3 & 210.6 & 220.8 & 93.3  & 139.1 & 267.7 & 169.5 & 333.8 \\
 & GradAscent+    & 237.6 & 163.8 & 243.2 & 274.5 & 95.9  & 145.8 & 336.2 & 170.2 & 336.1 \\
 & NegTaskVector+ & 204.7 & 130.4 & 188.1 & 211.1 & 86.4  & 127.5 & 265.0 & 170.1 & 325.3 \\
 & \textsc{LingTea} (ours) & 206.4 & 134.1 & 195.1 & 241.0 & 91.2  & 151.1 & 317.3 & 189.1 & 418.4 \\
\multirow{-5}{*}{BLOOM-560M} &
  \cellcolor[HTML]{EFEFEF}Oracle &
  \cellcolor[HTML]{EFEFEF}204.4 &
  \cellcolor[HTML]{EFEFEF}101.3 &
  \cellcolor[HTML]{EFEFEF}139.5 &
  \cellcolor[HTML]{EFEFEF}163.4 &
  \cellcolor[HTML]{EFEFEF}84.2 &
  \cellcolor[HTML]{EFEFEF}509.5 &
  \cellcolor[HTML]{EFEFEF}237.1 &
  \cellcolor[HTML]{EFEFEF}164.2 &
  \cellcolor[HTML]{EFEFEF}393.9 \\ \hline
 & Original       & 89.5  & 86.4  & 92.6  & 105.9 & 47.1  & 63.9  & 98.5  & 83.7  & 117.5 \\
 & GradAscent+    & 258.6 & 208.3 & 218.2 & 228.3 & 77.4  & 108.9 & 235.2 & 116.6 & 169.6 \\
 & NegTaskVector+ & 104.4 & 106.4 & 109.3 & 128.8 & 51.1  & 71.2  & 120.6 & 101.8 & 137.2 \\
 & \textsc{LingTea} (ours) & 137.0 & 84.6  & 118.4 & 107.3 & 56.0  & 86.3  & 114.3 & 119.2 & 297.3 \\
\multirow{-5}{*}{BLOOM-3B} &
  \cellcolor[HTML]{EFEFEF}Oracle &
  \cellcolor[HTML]{EFEFEF}136.5 &
  \cellcolor[HTML]{EFEFEF}65.7 &
  \cellcolor[HTML]{EFEFEF}65.1 &
  \cellcolor[HTML]{EFEFEF}81.9 &
  \cellcolor[HTML]{EFEFEF}37.6 &
  \cellcolor[HTML]{EFEFEF}67.8 &
  \cellcolor[HTML]{EFEFEF}74.8 &
  \cellcolor[HTML]{EFEFEF}150.5 &
  \cellcolor[HTML]{EFEFEF}193.5 \\ \Xhline{2\arrayrulewidth}
\end{tabular}
}
\caption{Perplexity of Test Set in BMLAMA-53.}
\label{tab:bmlama_test_ppl}
\end{table*}

\end{document}